\pdfoutput=1

\documentclass[11pt]{article}

\usepackage{amsmath,amsfonts,bm}

\def\eqref#1{equation~\ref{#1}}

\def\1{\bm{1}}

\DeclareMathAlphabet{\mathsfit}{\encodingdefault}{\sfdefault}{m}{sl}
\SetMathAlphabet{\mathsfit}{bold}{\encodingdefault}{\sfdefault}{bx}{n}

\usepackage[preprint]{acl}

\usepackage{times}
\usepackage{latexsym}

\usepackage[T1]{fontenc}

\usepackage[utf8]{inputenc}

\usepackage{microtype}

\usepackage{inconsolata}

\usepackage{graphicx}

\usepackage{amsmath}
\usepackage{graphicx}

\usepackage{booktabs}
\usepackage{multirow}

\usepackage{graphicx}
\usepackage{fancyhdr}
\usepackage{tikz}
\usepackage{mathrsfs}
\usepackage{color}
\usepackage{multirow}
\usepackage{diagbox}
\usepackage{pifont}
\usepackage{mathrsfs}
\usepackage{threeparttable}
\usepackage{comment}
\usepackage[linesnumbered,ruled,vlined]{algorithm2e}
\usepackage{amsthm}
\usepackage{enumitem}
\usepackage[noabbrev,capitalise]{cleveref}
\usepackage{footnote}
\usepackage{subcaption}
\usepackage{cleveref} 
\usepackage{makecell}
\usepackage{listings}  
\makesavenoteenv{tabular}
\makesavenoteenv{table}
\usepackage{wrapfig}
\usepackage{listings}
\usepackage[toc,page]{appendix}
\usepackage{float}

\theoremstyle{plain}

\theoremstyle{definition}

\theoremstyle{remark}

\usepackage[textsize=tiny]{todonotes}
\usepackage{listings} 

\usepackage{pifont}
\usepackage{xspace}
\newcommand{\methodFullName}{Self-Adjusting Softmax\xspace}
\newcommand{\methodShortName}{SA-Softmax\xspace}

\title{Self-Adjust Softmax}

\author{
 \textbf{Chuanyang Zheng\textsuperscript{1}},
 \textbf{Yihang Gao\textsuperscript{2}},
 \textbf{ Guoxuan Chen\textsuperscript{3}},
 \textbf{Han Shi\textsuperscript{4}},
\\
 \textbf{ Jing Xiong\textsuperscript{3}},
 \textbf{Xiaozhe Ren\textsuperscript{4}},
 \textbf{Chao Huang\textsuperscript{3}},
 \textbf{Xin Jiang \textsuperscript{4}},
\\
 \textbf{ Zhenguo Li\textsuperscript{4}},
 \textbf{ Yu Li\textsuperscript{1}},
\\
\\
 \textsuperscript{1}The Chinese University of Hong Kong,
 \textsuperscript{2}National University of Singapore\\
 \textsuperscript{3}The University of Hong Kong,
 \textsuperscript{4}Noah's Ark Lab
\\
 \small{
 \textbf{Code:} \href{https://github.com/chuanyang-Zheng/SA-Softmax}{https://github.com/chuanyang-Zheng/SA-Softmax}} \\
   \small{ \textbf{Contact:} \href{cyzheng21@link.cuhk.edu.hk}{cyzheng21@link.cuhk.edu.hk}
 }
}

\begin{document}
\maketitle
\begin{abstract}
The softmax function is crucial in Transformer attention, 
which normalizes each row of the attention scores with summation to one, achieving superior performances over other alternative functions.
However, the softmax function can face a gradient vanishing issue when some elements of the attention scores approach extreme values, such as probabilities close to one or zero.
In this paper, we propose \methodFullName (\methodShortName) to address this issue by modifying $softmax(x)$ to $x \cdot softmax(x)$ and its normalized variant $\frac{(x - min(x_{\min},0))}{max(0,x_{max})-min(x_{min},0)} \cdot softmax(x)$.
We theoretically show that \methodShortName provides enhanced gradient properties compared to the vanilla softmax function.
Moreover, \methodShortName Attention can be seamlessly integrated into existing Transformer models to their attention mechanisms with minor adjustments.
We conducted experiments to evaluate the empirical performance of Transformer models using \methodShortName compared to the vanilla softmax function. 
These experiments, involving models with up to 2.7 billion parameters, are conducted across diverse datasets, language tasks, and positional encoding methods.
\end{abstract}

\section{Introduction}
Transformer-based models \citep{vaswani2017attention} have delivered exceptional performances across widespread applications, including language processing \citep{zhang2020pegasus,guo2021longt5,ainslie2023colt5}, computer vision \citep{alexey2020image,touvron2021training,liu2021swin,chen2024pixartalpha,peebles2023scalable}, quantitative research \citep{zhou2024dont,liu2021finbert,wu2023bloomberggpt}, and scientific machine learning \citep{taylor2022galactica,geneva2022Transformers}. 
A critical component of the Transformer is its attention mechanism, 
which computes the importance and contribution of each token in a sequence for next-token generation.
Central to this mechanism is the softmax function, a mathematical operation that normalizes attention scores 
token-wise, ensuring a summation of one.
This property facilitates probabilistic interpretability and enables a more expressive attention mechanism. For example, \citet{chen2024sepllm,xiao2024infllm} observed that most attention scores are usually concentrated on specific tokens, allowing for more efficient Transformer architectures by discarding tokens with lower accumulative attention scores. As a result, the normalized attention scores produced by softmax provide insights into the mechanism of next-token generation in LLMs. Moreover, compared to other attention functions, softmax exhibits some unique and advantageous properties, which contribute to the superior performance of softmax-based Transformer models~\citep{han2024bridging,deng2023superiority}.
One of the primary limitations of softmax lies in its susceptibility to the gradient vanishing problem. When input values to the softmax function become highly polarized, i.e., extreme values that are very large or small, the resulting probabilities can exhibit extreme sparsity. This, in turn, leads to gradients that approach zero, impeding effective learning and optimization during backpropagation. Such issues are particularly pronounced in deep architectures, where the accumulation of small gradients can hinder convergence and degrade model performance \cite{vaswani2017attention,duvvuri2024laser}.
Several variations have been proposed, including ReLU attention \cite{nair2010rectified,chen2020arelu,wortsman2023replacing,shen2023study} or sigmoid attention \cite{ramapuram2024theory}. These alternatives aim to address specific shortcomings of softmax, such as its sensitivity to extreme input values or its restricted output range, which may limit the behavior of the attention mechanism. However, these approaches often fall short of achieving comparable stability, interpretability, or general performance, especially in large-scale models where softmax continues to dominate due to its robustness and simplicity.


To address this limitation, we propose a novel modification to the softmax function, introducing \methodFullName (\methodShortName), which enhances gradient propagation while preserving the probabilistic properties and ranking order of traditional softmax. 
Our approach builds on theoretical insights and empirical observations. First, we show theoretically that modifying the softmax function to $x \cdot softmax(x)$ amplifies gradient magnitudes, addressing gradient saturation under a range of typical conditions. Building on this formulation, we further refine the formulation to $\frac{(x - min(x_{\min},0))}{max(0,x_{max})-min(x_{min},0)} \cdot softmax(x)$, incorporating the normalization while enhancing gradient flow. It also maintains the relative ordering of input values, which serves as a critical property for the effectiveness of attention mechanisms. The proposed modification of the vanilla softmax function ensures compatibility with standard Transformer architectures and facilitates seamless integration into existing frameworks.
\begin{enumerate} 
    \item We propose $x \cdot softmax(x)$ as an alternative to the vanilla softmax in the attention mechanism to improve gradient magnitudes, thereby enhancing backpropagation during training.
    Additionally, we refine $x \cdot softmax(x)$ to $\frac{(x - min(x_{\min},0))}{max(0,x_{max})-min(x_{min},0)} \cdot softmax(x)$ with normalization, preserving a critical property of softmax while achieving superior performance. 
    
    \item We conduct extensive experiments across various datasets, tasks, and models, comparing the proposed \methodShortName and its variants with the standard $softmax(x)$. Results demonstrate that our approach effectively mitigates gradient vanishing and consistently improves performances across models with different scales. 
    \item We validate the proposed methods on large-scale pre-training datasets with a training length of 2048. Moreover, we also show the effectiveness of the proposed method in downstream tasks.
    \end{enumerate}

\section{Related Works}
\paragraph{Transformer Attention.}
The Transformer model, introduced by Vaswani et al. \cite{vaswani2017attention}, revolutionized the field of Natural Language Processing (NLP) with its self-attention mechanism. Unlike previous sequence models such as RNNs and LSTMs \cite{graves2012long}, Transformer does not rely on recurrent structures and instead uses self-attention to depict relationships between input tokens in parallel.
Self-attention, also known as scaled dot-product attention, computes attention scores between input tokens using the query ($Q$), key ($K$), and value ($V$) vectors. 
\[
\text{Attention}(Q, K, V) = softmax\left(\frac{QK^T}{\sqrt{d_k}}\right)V
\]
where $d_k$ is the dimension of the key vectors \cite{vaswani2017attention}.
There are also \textit{linearized attention} methods, such as the Linformer \cite{wang2020linformer} and Performer \cite{choromanski2020rethinking}, approximating the softmax attention function using low-rank approximations, reducing the computational complexity from $O(n^2)$ to $O(n)$.
Another approach to reduce computational complexity is through \textit{sparse attention}, where only a subset of attention scores are computed. For example, the Longformer \cite{beltagy2020longformer} uses a combination of local windowed attention and global attention, reducing the attention complexity to $O(n)$ for sequences of length $n$.

\paragraph{Gradient Vanishing.} The gradient vanishing problem refers to the phenomenon where gradients become exceedingly small during backpropagation \cite{lillicrap2020backpropagation}.
Several works have explored the causes and potential solutions to the gradient vanishing problem. Gradient clipping \cite{zhang2019gradient} is one practical solution to mitigate both vanishing and exploding gradients. This technique caps gradients at a maximum value to prevent them from becoming too small or too large. \citet{pascanu2013difficulty} explored gradient clipping in the context of RNNs and found that it can help stabilize training by preventing gradient explosions, which often arise due to large gradients propagating backward through deep networks. The skip connection \cite{he2016deep} is the potential way to mitigate the gradient vanishing problem. For softmax attention, the gradient will become zero if one attention probability is too large \cite{vaswani2017attention}.

\paragraph{Normalization.} Batch Normalization (BN) \cite{ioffe2015batch} normalizes activations along the batch dimension, while Layer Normalization (LN) \cite{ba2016layer} operates along the channel dimension, and Instance Normalization (IN) \cite{huang2017arbitrary} applies BN-like computations independently for each sample. Weight Normalization (WN) \cite{salimans2016weight} instead normalizes filter weights directly. Group Normalization (GN) divides channels into groups, normalizing each group independently, and its computations are unaffected by batch size. \citet{bjorck2018understanding} show that in networks without BN, large gradient updates can cause diverging loss and uncontrolled activation growth with network depth, limiting learning rates. Similarly, \citet{xu2019understanding} demonstrates that layer normalization smooths gradients and highlights the importance of mean and variance derivatives, which re-center and re-scale backward gradients beyond forward normalization.

\section{Method}

\subsection{Softmax Attention Mechanism}
In the attention mechanism, the weight \( \alpha_{ij} \) represents the attention score between token $i$ (the query) and token $j$ (the key). This score quantifies the relative importance of token $j$ to token $i$, among all tokens in the input sequence. It is formulated as

\begin{equation}
\alpha_{ij} = softmax\left(\frac{q_i^T k_j}{\sqrt{d_k}}\right) = \frac{\exp\left(\frac{q_i^T k_j}{\sqrt{d_k}}\right)}{\sum_{j'} \exp\left(\frac{q_i^T k_{j'}}{\sqrt{d_k}}\right)},
\end{equation}
where \( q_i \) and \( k_j \) are the query and key vectors for tokens $i$ and $j$, respectively, and \( d_k \) is a scaling factor based on the dimensionality of the keys \cite{vaswani2017attention}. 
The softmax function ensures that the resulting attention scores \( \alpha_{ij} \) are normalized and can be interpreted as probabilities, summing to one over all tokens $j$ for a given query token $i$.

The final output of the attention mechanism for each query token \( i \) is then calculated as a weighted sum of the values \( v_j \) corresponding to each token $j$ in the sequence, with the weight determined by the attention scores \( \alpha_{ij} \). The output of the attention mechanism for token $i$ is defined as
\begin{equation}
\text{Attention}_{i}(Q, K, V) = \sum_j \alpha_{ij} v_j,
\end{equation}
where \( Q \), \( K \), and \( V \) are matrices representing all queries, keys, and values for a given sequence. This approach allows the model to focus selectively on parts of the sequence that contribute meaningfully to the current query position \cite{bahdanau2014neural}. 

\subsection{Gradient of Softmax Attention}
Training Transformer models involves updating all trainable parameters using their gradients. The backpropagation process, which relies on the chain rule, requires the computation of the derivative of the softmax function with respect to its inputs. However, when the input values to the softmax function become extremely large or small, the function can enter flat regions. This results in vanishing gradients, which can hinder the efficient training of model parameters.


We denote the pre-softmax attention scores (i.e., the input to the softmax function before normalization) as
\begin{equation}
z_{i,j} = \frac{q_i^T k_j}{\sqrt{d_k}},
\end{equation}
then the derivative of the output attention scores (after passing through the softmax function) with respect to the input \( z_{i,j} \) admits
\begin{equation}
\label{eq_derivatives}
    \begin{split}
        & \frac{\partial \alpha_{ij}}{\partial z_{i,j}} = \alpha_{ij} (1 - \alpha_{ij}),\\
        & \frac{\partial \alpha_{ij}}{\partial z_{i,j^{\prime}}} = -\alpha_{ij} \alpha_{ij^{\prime}}, \quad \text{for } j^{\prime} \neq j.
    \end{split}
\end{equation}

This Jacobian matrix structure implies that each attention weight depends not only on its own value but also on the values of all other weights. This property, while beneficial for capturing complex relationships, can also make optimization challenging in some scenarios, as explored in the next section.

\subsection{Gradient Vanishing in Softmax Attention}

One notable issue with softmax attention is the vanishing gradient problem, especially when attention scores become highly peaked. When the softmax output approaches 1 for a specific score and 0 for others, the gradients can become excessively small, slowing down or even halting learning. This is particularly problematic in deeper models where multiple layers of attention are stacked.

The vanishing gradient issue arises from the form of the softmax derivatives. Let us examine the two cases:
Consider the token $i$ in the attention mechanism,
and let $z_{i,j}$ and $\alpha_{i,j}$ represent the attention scores of all tokens relative to token $i$, for $j = 1,2,\ldots,T$. In the extreme case where one of the attention weights dominates, i.e., $\alpha_{i,j^{*}} \approx 1$ and $\alpha_{i,j} \approx 0$, for $j \neq j^{*}$. Then Equation \ref{eq_derivatives} implies that $\frac{\partial \alpha_{i,j}}{\partial z_{i,j^{\prime}}} \approx 0$ for all $j, j^{\prime}=1,2,\ldots, T$.
This result indicates that, under such circumstances, the derivative of the output attention weights with respect to the input pre-softmax attention scores vanishes, leading to gradient vanishing across all tokens.
Moreover, in a milder case where $\alpha_{i,j} \approx 0$ holds for some $j$, we have $\frac{\partial \alpha_{i,j}}{\partial z_{i,j^{\prime}}} \approx 0$ and $\frac{\partial \alpha_{i,j^{\prime}}}{\partial z_{i,j}} \approx 0$ for $j^{\prime} = 1,2,\ldots,T$. This means that the derivative of the softmax function partially vanishes, if input and output correspond to $\alpha_{i,j}$ and $z_{i,j}$ of token $j$, resulting in gradient vanishing for those specific tokens.

In summary, when the extreme case arises where one attention score dominates while others approach zero, the softmax mechanism suffers from complete gradient vanishing for all tokens, leading to slow training and failure in gradient backpropagation. In the milder case, where some attention scores are close to zero, the derivatives associated with these tokens and their attention scores still vanish, causing suboptimal training performance.

The extreme case, where some attention scores approach zero, frequently occurs in attention mechanisms due to the exponential function's sensitivity to large values. In the following section, we first introduce a modification to the vanilla softmax, called \methodFullName (\methodShortName), which is theoretically guaranteed to enhance and amplify gradient propagation. Additionally, we propose several variants of \methodShortName, designed to further improve the effectiveness and stability of Transformer models by incorporating normalization techniques.

\subsection{\methodFullName}
To address the issue of potential gradient vanishing of the softmax function, we propose modifying the attention mechanism by scaling the softmax output with its input, called \methodFullName (\methodShortName). Specifically, we redefine the output of attention scores as follows:
\begin{equation}
\beta_{i,j} = z_{i,j} \cdot softmax(z_{i,j}),
\end{equation}
where \( z_{i,j} \) is the pre-softmax attention score of the token $i$ corresponding to the token $j$. This modification introduces an additional scaling term \( z_{i,j} \) to calculate the final attention scores besides the standard softmax function, amplifying the gradient propagation compared to the original formulation.

\paragraph{Gradient Analysis of \methodShortName.}
Let us evaluate the gradient of the modified attention scores \( \beta_{i,j} \) with respect to the input \( z_{i,j^{\prime}} \). Differentiating \( \beta_{i,j} = z_{i,j} \cdot softmax(z_{i,j}) \) with respect to \( z_{i,j^{\prime}} \), we have

\begin{equation}
\label{eq:sa-softmax i=j}
\begin{split}
    \frac{\partial \beta_{i,j}}{\partial z_{i,j}} & = softmax(z_{i,j}) + z_{i,j} \cdot \frac{\partial softmax(z_{i,j})}{\partial z_{i,j}}\\
    & = \alpha_{i,j} + z_{i,j} \cdot \alpha_{i,j} (1- \alpha_{i,j}),
\end{split}
\end{equation}
and
\begin{equation}
\label{eq:sa-softmax i/=j}
    \begin{split}
        \frac{\partial \beta_{i,j}}{\partial z_{i,j^{\prime}}} & = z_{i,j^{\prime}} \cdot \frac{\partial softmax(z_{i,j})}{\partial z_{i,j^{\prime}}}\\
        & = - z_{i,j} \cdot \alpha_{i,j} \alpha_{i,j^{\prime}},
    \end{split}
\end{equation}
with $j^{\prime} \neq j$.

\paragraph{Implications for Gradient Vanishing.}
According to Equation \ref{eq:sa-softmax i=j}, considering the extreme case where $\alpha_{i,j^{*}} \approx 1$, the gradient is amplified as the first term $\alpha_{i,j^{*}}$ is dominant and governs the gradient. Moreover, for tokens where $\alpha_{i,j} \approx 0$, the gradient is enhanced by the dynamic and self-adjusting scaler $z_{i,j}$, as demonstrated in Equations \ref{eq:sa-softmax i=j} and \ref{eq:sa-softmax i/=j}. Therefore, our method significantly enhances the gradient propagation for tokens with $\alpha_{i,j^{\prime}} \approx 1$ and improves the gradient for tokens satisfying $\alpha_{i,j} \approx 0$ through the dynamic and self-adjusting scalers.
\paragraph{ Comparison with Standard Softmax.}
In the standard softmax attention, the gradient softmax output \( \alpha_{i,j} \) with respect to the input \( z_{i,j} \) tends to vanish when \( \alpha_{i,j} \) approaches 0 or 1. By introducing an additional self-adjusting term in the attention score computation (i.e., modufying $softmax(x)$ to \( x \cdot softmax(x) \)), we allow for a more resilient gradient. As shown in Equations \ref{eq:sa-softmax i=j} and \ref{eq:sa-softmax i/=j}, this approach may not completely eliminate the gradient vanishing problem, it significantly mitigates its effects, especially in cases with long sequences or deep networks, where gradients from softmax attention typically diminish~\cite{vaswani2017attention}.

\subsection{Variants of \methodShortName}
In this section, we further develop some variants of \methodShortName, by utilizing normalization techniques on the self-adjusting term, to further stablize the training process. 

\paragraph{Variant 1: \( (x - x_{\min}) \cdot softmax(x) \)}
A notable potential inconsistency in \methodShortName arises from the negative attention scores, which can lead to unpredictable and difficult-to-interpret behavior in the attention mechanism.

To address this issue, we propose a modified approach that shifts the self-adjusting term by its minimum value along the sequence. Specifically, we reformulate the attention computation as \( (x - x_{\min}) \cdot softmax(x) \), where \( x_{\min} \) represents the minimum value of \( x \) across the sequence. This modification ensures that all attention scores are non-negative, thereby stabilizing the scaling effect across different \( x_i \):
\begin{equation}
\gamma_{i,j} = (z_{i,j} - z_{i,\min}) \cdot softmax(z_{i,j})
\end{equation}
where $z_{i,\min}:=\min\{z_{i,j}: j=1,2,\ldots,T\}$ denotes the minimum values of $z_{i,j}$ along the sequence.
This adjustment enhances the robustness of the attention mechanism by ensuring consistency and stability in the scaling of attention scores.

\paragraph{Variant 2: \( \frac{x - x_{\min}}{x_{\max}-x_{\min}} \cdot softmax(x) \)}
The first variant introduced a shift in the self-adjusting term to ensure the non-negativity of attention scores. Building on this idea, a more widely used technique is normalization. To further stabilize training, we normalize the self-adjusting term to $ \frac{x - x_{\min}}{x_{\max}-x_{\min}} \in [0,1]$. Therefore, the attention scores are calculated as follows:
\begin{equation}
\delta_{i,j} = \frac{z_{i,j} - z_{i,\min}}{z_{i,\max} - z_{i,\min}} \cdot softmax(z_{i,j}),
\end{equation}
where \( z_{i,\max} = \max\{z_{i,j}:j=1,2,\ldots,T\} \) denotes the maximum value of $z_{i,j}$ along the sequence, and $z_{i,\min}:=\min\{z_{i,j}: j=1,2,\ldots,T\}$ represents the minimum value. This normalization ensures that the self-adjusting term \( \frac{(x - x_{\min})}{x_{\max} - x_{\min}} \) lies within the bounded region \([0, 1]\), resulting in a stable scaling effect across different input distributions.
This formulation provides a stable gradient computation, as the adjusting term, normalized by \( x_{\max} - x_{\min} \), prevents excessively large values and ensures all values falling within a bounded range.



\paragraph{Variant 3: \( \frac{x - min(x_{\min},0)}{max(0,x_{\max})-min(x_{\min},0)} \cdot softmax(x) \)}
To make it easier for the model optimization, we add a threshold to further normalize the  $ \frac{(x - x_{\min})}{x_{max}-x_{min}}$ $x - x_{\min}$ to $ \frac{x - min(x_{\min},0)}{max(0,x_{\max})-min(x_{\min},0)} \in$ [0,1].
\begin{equation}
\delta_i = \frac{x - min(x_{\min},0)}{max(0,x_{\max})-min(x_{\min},0)} \cdot softmax(x_i)
\end{equation}
where \( x_{\max} = \max(x) \) and \( x_{\min} = \min(x) \). When the $x$ becomes positive and \( x_{\max} - x_{\min} \gg 0 \)  ,  $\frac{x - min(x_{\min},0)}{max(0,x_{\max})-min(x_{\min},0)}$ will close to 1 so that the $\frac{x - min(x_{\min},0)}{max(0,x_{\max})-min(x_{\min},0)} \cdot softmax(x_i)$ degrades to $softma(x)$.

\section{Experiment}
\paragraph{Datasets.} We use the Arxiv and Books3 dataset for our experiments. The training is conducted using a batch size of 512 or 1024 sequences, each with a sequence length from length 128 to length 2048. The models are trained for 50000 iterations. Throughout the training process, we monitor both the training and gradient. We also evaluate our methods in downstream datasets, such as sequence classification and machine translation.

\paragraph{Experiment Setting.} We begin by conducting experiments on the Arxiv and Books datasets, evaluating model performance across training sequence lengths ranging from 128 to 1024 with various positional encodings. Next, we validate our method on models of varying scales, from 125M to 2.7B parameters. Following this, we analyze the performance of different model variants and assess their ability to extrapolate to longer sequence lengths. Subsequently, we further validate the method on downstream tasks, including text classification and machine translation. Lastly, we visualize the gradient behavior across different methods to provide deeper insights into their effectiveness. The experiment setting details are presented in Appendix \ref{appendix: experiment setting}. By default, we use the \( \frac{x - min(x_{\min},0)}{max(0,x_{\max})-min(x_{\min},0)} \cdot softmax(x) \).

\subsection{Compare with Baseline Performance}
\begin{table}[htb]
\caption{The perplexity on Arxiv and Books dataset with different position encodings.}
\centering
\resizebox{0.49\textwidth}{!}{
\begin{tabular}{cccccc}
\toprule
\textbf{Data}&\textbf{PE}&\textbf{\methodShortName} & 128 & 512 & 1024  \\ \midrule
Arxiv&Kerple& \ding{53} &   14.61&6.70 & 5.47\\ 
Arxiv&Kerple& \ding{51} & 14.51 & 6.66& 5.44 \\ 
Arxiv&FIRE& \ding{53} &   14.76&6.67 & 5.43\\ 
Arxiv&FIRE& \ding{51} & 14.46 & 6.59& 5.38 \\ 
Arxiv&RoPE& \ding{53} &   14.86&6.70 & 5.52\\ 
Arxiv&RoPE& \ding{51} & 14.62 & 6.63& 5.49 \\ 
Arxiv&DAPEV2-Kerple& \ding{53} &   14.27&6.63 & 5.26\\ 
Arxiv&DAPEV2-Kerple& \ding{51} & \textbf{14.10} & \textbf{6.36}& \textbf{5.20} \\ 
\midrule
Books&Kerple &\ding{53} &88.88 & 38.46 & 28.65 \\  
Books&Kerple& \ding{51} &87.56 & 37.95 & 28.37\\ 
Books&FIRE &\ding{53} &88.48 & 38.12 & 28.57 \\  
Books&FIRE& \ding{51} &87.98 & 37.34 & 28.00\\ 
Books&RoPE &\ding{53} &89.60 & 38.29 & 30.53 \\  
Books&RoPE& \ding{51} &89.36 & 37.57 & 30.29\\ 
Books&DAPEV2-Kerple &\ding{53} &84.33 & 36.25 & 26.86\\  
Books&DAPEV2-Kerple &\ding{51} &\textbf{83.63}&\textbf{35.93} &\textbf{26.56} \\  
\bottomrule
\end{tabular}
}
\label{table: different_pe}
\end{table} 

\paragraph{The \methodShortName could improve performance improvements across different position encodings.}
The results in Table \ref{table: different_pe} highlight the effectiveness of \methodShortName in improving perplexity for Kerple, FIRE, RoPE and DAPEV2-Kerple. Without \methodShortName (\ding{53}), RoPE achieves perplexities of 89.60, 38.29, and 30.53 for sequence lengths 128, 512, and 1024, respectively. With \methodShortName (\ding{51}), these values drop to 89.36, 37.57, and 30.29, showcasing its contribution. DAPEV2-Kerple exhibits more significant improvements, with perplexities dropping from 84.33, 36.25, and 26.86 to 83.63, 35.93, and 26.56 across the respective sequence lengths when \methodShortName is applied. This demonstrates the universal applicability of \methodShortName to enhance position encoding methods.

\paragraph{DAPEV2-Kerple achieves the best performance, especially with \methodShortName.}
Among all tested configurations, DAPEV2-Kerple combined with \methodShortName yields the lowest perplexity scores, outperforming both the baseline RoPE and RoPE with \methodShortName. For instance, at a sequence length of 1024, DAPEV2-Kerple with \methodShortName achieves a perplexity of 26.56, compared to 30.29 for RoPE with \methodShortName. This superiority is consistent across shorter sequence lengths as well, with DAPEV2-Kerple maintaining its advantage even without \methodShortName. These results confirm that DAPEV2-Kerple is the most effective position encoding method for reducing perplexity in language modeling tasks.

\paragraph{The proposed \methodShortName improves both short and long-sequence modeling.}
The analysis indicates that \methodShortName enhances performance at all sequence lengths, demonstrating its ability to handle both short-range and long-range dependencies effectively. The reductions in perplexity are keeped at longer sequence lengths, particularly for DAPEV2-Kerple (e.g., a drop from 26.86 to 26.56 for length 1024), suggesting that \methodShortName still provides better optimization for long contexts. This capability is critical for modern language models that often deal with extensive input sequences.

\begin{table}[!htbp]
\caption{The perplexity on Arxiv and  Books dataset with training length 2048, evaluate from length 128 to length 2048.}
\centering
\resizebox{0.49\textwidth}{!}{
\begin{tabular}{cccccccc}
\toprule
\text{Dataset}&PE&\methodShortName&  128 & 256&512&1024&2048  \\ \midrule
Arxiv &RoPE&\ding{53} &9.14&7.53 & 5.42&5.00&4.95\\ 
Arxiv &RoPE&\ding{51}&9.05 &7.46 &5.38&4.96&4.92\\ 
Arxiv &DAPEV2-Kerple&\ding{53} &8.80&7.22 & 5.16&4.74&4.64\\ 
Arxiv &DAPEV2-Kerple&\ding{51}&8.70 &7.15 &5.13&4.72&4.61\\ \midrule
Books &RoPE&\ding{53} &35.99&31.32 & 25.97&24.32&22.65\\ 
Books &RoPE&\ding{51}&35.71 &31.15 &25.906&24.23&22.63\\ 
Books &DAPEV2-Kerple&\ding{53} &34.13&29.54 & 24.28&22.60&20.85\\ 
Books &DAPEV2-Kerple&\ding{51}&33.60 &29.16 &24.06&22.40&20.71\\ \midrule 
\end{tabular}
}
\label{table:lengh 2048}
\end{table}

\paragraph{The \methodShortName still works well on longer training length.} The results in Table \ref{table:lengh 2048} demonstrate the impact of using \methodShortName (SA-Softmax, indicated by \ding{51}) versus not using it (\ding{53}) on the Arxiv and Books datasets under different positional encodings (RoPE and DAPEV2-Kerple) across evaluation lengths from 128 to 2048, with a training length of 2048. For both datasets and positional encodings, \methodShortName consistently improves performance, as evidenced by lower perplexity values. On the Arxiv dataset, DAPEV2-Kerple with \methodShortName achieves the best results, with perplexity decreasing from 8.70 at length 128 to 4.61 at length 2048, outperforming baseline DAPEV2-Kerple in all cases. For the Books dataset, DAPEV2-Kerple combined with \methodShortName achieves the lowest perplexity. Similarly, RoPE with \methodShortName also achieves better performance than baseline RoPE on Arxiv and Books dataset from evaluation length 128 to length 2048. These results indicate that \methodShortName effectively enhances model performance, and works well on longer training length.


\subsection{The performance on Larger Model Size}
\begin{table}[htb]
\caption{The perplexity on Arxiv and Books dataset with different model sizes, with training length 512.}
\centering
\resizebox{0.5\textwidth}{!}{
\begin{tabular}{ccccccc}
\toprule
\textbf{PE}&\textbf{Dataset}&\textbf{\methodShortName}&125M & 350M & 1.3B & 2.7B  \\ 
\midrule
RoPE&Arxiv& \ding{53}&6.70 & 6.26 & 6.01 &5.93\\
RoPE&Arxiv&\ding{51} &6.63 &6.20 &5.92 & 5.83 \\ \midrule
DAPEV2-Kerple&Arxiv&\ding{53}&6.63 & 6.02 & 5.79 &5.70\\
DAPEV2-Kerple&Arxiv&\ding{51} &6.36 &5.97 &5.74&5.65 \\ \midrule
RoPE&Book & \ding{53} &38.29 &33.81 & 30.94 &29.98 \\
RoPE&Book&   \ding{51} &37.57 & 33.17 &30.24 & 29.15 \\ \midrule
DAPEV2-Kerple&Book&\ding{53} &36.25 &32.20 & 29.32 &28.15 \\
DAPEV2-Kerple&Book& \ding{51} &35.93 & 31.82 &28.91 & 27.75 \\
\bottomrule
\end{tabular}
}
\label{table:large_model}
\end{table}

\paragraph{\methodShortName still enhances performance for larger model sizes.}
Table \ref{table:large_model} demonstrates the effectiveness of \methodShortName across different model sizes, ranging from 125M to 2.7B parameters, on the Books and Arxiv datasets. The results show that, as model size increases, the integration of \methodShortName consistently improves performance compared to the baseline (\ding{53}). For example, with RoPE on the Books dataset, the perplexity at 2.7B parameters decreases from 29.98 to 29.15 when \methodShortName is applied. Similarly, for DAPEV2-Kerple on the same dataset, perplexity improves from 28.15 to 27.75 at the largest model size, highlighting the compatibility of \methodShortName with large-scale models.


\paragraph{\methodShortName delivers consistent improvements across datasets.}
The results are consistent across both the Books and Arxiv datasets, confirming the generalizability of \methodShortName. On the Arxiv dataset, for instance, RoPE with \methodShortName reduces perplexity across all model sizes, from 6.70 to 6.63 at 125M parameters and from 5.93 to 5.83 at 2.7B parameters. Similar trends are observed for DAPEV2-Kerple, where the improvements are slightly less pronounced but still consistent. These findings indicate that \methodShortName is robust and effective across diverse text corpora and model configurations.

\subsection{The Performance of Different Variants}
\begin{table}[htb]
\caption{The perplexity on the Books dataset with training length 512, compared to baselines.}
\centering
\resizebox{0.49\textwidth}{!}{
\begin{tabular}{ccccc}
\toprule
\text{Length}&\textbf{Variant}&  RoPE & $DAPEV2-Kerple$  \\ \midrule
128 &$softmax(x)$ &89.60 & 84.33&\\ 
128 &$x*softmax(x)$ &85.98 &82.63 &\\  
128 &$(x-x_{max})*softmax(x)$ &86.10 &\textbf{82.08} &\\ 
128 &$\frac{(x - x_{\min})}{x_{max}-x_{min}} \cdot softmax(x)$ &89.36 &84.21 &\\  
128 &$\frac{(x - min(x_{\min},0))}{max(0,x_{max})-min(x_{min},0)} \cdot softmax(x)$ &\textbf{89.36} &83.63 &\\ 
\midrule
512 & $softmax(x)$&38.29 &36.25 &\\ 
512 & $x*softmax(x)$&38.07 &35.82 &\\  
512 &$(x-x_{max})*softmax(x)$ &39.47 &\textbf{35.70} &\\ 
512 &$\frac{(x - x_{\min})}{x_{max}-x_{min}} \cdot softmax(x)$ &37.93 &36.21 &\\ 
512 &$\frac{(x - min(x_{\min},0))}{max(0,x_{max})-min(x_{min},0)} \cdot softmax(x)$ &\textbf{37.57} &35.93 &\\ 
\midrule
1024 &$softmax(x)$ &28.78 & 26.86&\\ 
1024 &$x*softmax(x)$ &28.96 & 26.62 &\\  
1024 &$(x-x_{max})*softmax(x)$ &30.26 & 26.78 &\\  
1024 &$\frac{(x - x_{\min})}{x_{max}-x_{min}} \cdot softmax(x)$ &28.73 & 26.74 &\\  
1024 &$\frac{(x - min(x_{\min},0))}{max(0,x_{max})-min(x_{min},0)} \cdot softmax(x)$ &\textbf{28.57} &\textbf{26.56} &\\  
\bottomrule
\end{tabular}
}
\label{table:different_variants}
\end{table}

\paragraph{The $\frac{(x - \min(x_{\min}, 0))}{\max(0, x_{\max}) - \min(x_{\min}, 0)} \cdot softmax(x)$ variant is a robust default choice.}  
Table~\ref{table:different_variants} shows that this variant consistently improves perplexity across all sequence lengths (128, 512, and 1024) and for both RoPE and DAPEV2-Kerple position encodings. For RoPE, this variant achieves the best performance at 128, 512, and 1024 lengths, and for DAPEV2-Kerple, it also achieves better performance than baseline from length 128 to length 1024. This suggests that the variant balances performance improvements across different configurations, making it a reliable choice when specific experimental conditions are not predefined.

\paragraph{Optimal \methodShortName variants depend on the experimental setup and position encoding method.}  
The results reveal that different variants perform best under different conditions. For example, the $(x - x_{\max}) \cdot softmax(x)$ variant achieves the lowest perplexity for DAPEV2-Kerple at lengths 128 (82.08) and 512 (35.70), outperforming all other configurations for these specific setups. Similarly, the standard $x \cdot softmax(x)$ variant shows competitive performance at 1024-length sequences, achieving 26.62 for DAPEV2-Kerple. These variations highlight that while certain formulations may work well across the board, optimal performance often depends on the interaction between the sequence length and the positional encoding technique.

\subsection{Performance on Downstream Tasks}
\begin{table}[htb]
\caption{The performance on downstream tasks, with 125M model size and 300B training tokens.}
\resizebox{0.49\textwidth}{!}{
\begin{tabular}{cccc}
\toprule
\textbf{Dataset}&Metrics& Softmax & \methodShortName\\ \midrule
Lambda&ppl$\downarrow$&21.63&\textbf{20.43}\\ 
WikiText&ppl$\downarrow$&27.57&\textbf{27.47}\\ 
ARCEasy&acc$\uparrow$&45.92&\textbf{47.52}\\ 
HellaSwag&acc$\uparrow$&30.34&\textbf{30.42}\\ 
PiQA& acc$\uparrow$&64.64&\textbf{64.69}\\
OpenBookQA&acc$\uparrow$&16.80&\textbf{18.00}\\ 
SciQ&acc$\uparrow$&76.80&\textbf{77.60}\\ 
Winogrande&acc$\uparrow$&51.54&\textbf{51.85}\\ 
\bottomrule
\end{tabular}
}
\label{table: downstream tasks}
\vspace{-5pt}
\end{table}
\paragraph{Pretrain Setting.} We pre-train a 125M model with 300B tokens from the Pile dataset and evaluate the performance on the downstream tasks \cite{black2022gpt}. Following the setting of previous works \cite{black2022gpt}, the training steps are 143000 with training length of 2048 and a global batch size 1024.
\paragraph{The \methodShortName achieve better performance than baseline softmax.} 
As shown in Table \ref{table: downstream tasks}, \methodShortName demonstrates superior performance compared to the baseline Softmax model across a wide range of tasks. The improvements are particularly notable in tasks requiring language modeling and reasoning. For instance, on the Lambda dataset \cite{paperno2016lambada}, \methodShortName achieves a significant reduction in perplexity (ppl) from 21.63 to 20.43. 
Similarly, on WikiText \cite{merity17pointer}, \methodShortName reduces perplexity from 27.57 to 27.47.
Additionally, it attains higher accuracy (acc) on several datasets, including ARCEasy \cite{clark2018arc}, HellaSwag \cite{zellers2019hellaswag}, PiQA\cite{clark2018arc}, OpenBookQA\cite{bisk2020piqa}, SciQ \cite{welbl2017crowdsourcing}, and Winogrande \cite{kocijan2019winogrande}. These results underscore the effectiveness of \methodShortName, even with a relatively small model size, when trained on a large-scale corpus. with potential for further improvements through increased model size and training data.


\section{The Performance on Classification and Translation Taks}
\begin{table}[htb]
\caption{Accuracy achieved on various downstream classification tasks. The \textbf{Improve $\Delta$} column shows the improvement in percentage points when using \methodShortName compared to Softmax.}
\resizebox{0.45\textwidth}{!}{
\begin{tabular}{cccc}
\toprule
\textbf{Dataset}&Softmax& \methodShortName & $\Delta$ \\ \midrule
AG-News&93.75&95.83&2.08\\ 
DBPedia&99.11&100&0.09\\ 
Yelp-Review&65.00&67.50&2.50\\ 
YahooNews&72.92&73.96&1.04\\ 
AmazomNews&62.50&68.75&6.25\\ 
\bottomrule
\end{tabular}
}
\label{table: classification_task}
\end{table}

\begin{table}[htb]
\caption{Performance comparison on IWSLT2017 machine translation tasks. Bold values indicate the best performance for each pair.}
\resizebox{0.45\textwidth}{!}{
\begin{tabular}{ccccccc}
\toprule
\textbf{Input}&\methodShortName&en&nl&de&it&ro \\ \midrule
en &\ding{53}&- &25.98 & 22.53 &24.08 & 21.98 \\
en &\ding{51}&- &\textbf{26.25} & \textbf{23.57} &\textbf{24.67} & \textbf{22.21} \\ \midrule
nl &\ding{53}&31.43 &- & 18.57 &15.89 & 14.67 \\
nl &\ding{51}&\textbf{32.10} &- &\textbf{19.21} & \textbf{16.14}  & \textbf{15.04} \\ \midrule
de &\ding{53}&26.83 &18.44 &-&14.55 & \textbf{13.72} \\
de &\ding{51}&\textbf{27.49} & \textbf{18.76} &-& \textbf{14.76}&13.57  \\ \midrule
it &\ding{53}&28.31 &15.50 &15.65 &- &15.77 \\
it &\ding{51}&\textbf{28.55} &\textbf{15.65} &\textbf{15.97} & -&\textbf{16.09} \\ \midrule
ro &\ding{53}&28.75 &15.42 & 15.72&18.27 & -\\
ro &\ding{51}&\textbf{29.21} &\textbf{16.71} &\textbf{16.11} &\textbf{18.54} & -\\ 
\bottomrule
\end{tabular}
}
\label{table: machine_translation}
\end{table}
\paragraph{The Performance on Classification and Translation Taks, the experiment setting is presented in Appendix \ref{appendix: experiment setting}.} The results across both classification and machine translation tasks indicate the consistent effectiveness of \methodShortName over traditional methods such as Softmax. On classification tasks (Table~\ref{table: classification_task}), \methodShortName achieves notable improvements across all datasets, with the highest improvement observed on the AmazonNews dataset (+6.25 percentage points) and significant gains on AG-News (+2.08), Yelp-Review (+2.50), and YahooNews (+1.04). 
On machine translation tasks (Table~\ref{table: machine_translation}), \methodShortName consistently outperforms baseline methods across multiple language pairs. 
For instance, when translating from English (EN) to other languages, \methodShortName achieves the highest accuracy gains, with improvements such as +0.27 (EN to NL) and +1.04 (EN to DE). 
Across both downstream classification and machine translation tasks, the improvements are consistently observed. 
These results collectively indicate that \methodShortName enhances both the accuracy and generalization capabilities of models, making it a promising alternative to traditional Softmax-based approaches. 



\subsection{Visualization of Attention Output}


We also visualize the attention probability for different methods in Appendix \ref{appendix: visualization attention}.

\paragraph{$\frac{(x - \min(x_{\min}, 0))}{\max(0, x_{\max}) - \min(x_{\min}, 0)} \cdot softmax(x)$ and $softmax(x)$ present similar pattern, compared to $\text{x} \cdot softmax(x)$}. As shown Appendix \ref{appendix: visualization attention}, the $\frac{(x - \min(x_{\min}, 0))}{\max(0, x_{\max}) - \min(x_{\min}, 0)} \cdot softmax(x)$ range may be larger than baseline $softmax(x)$. Also, the $softmax(x)$ and $\frac{(x - \min(x_{\min}, 0))}{\max(0, x_{\max}) - \min(x_{\min}, 0)} \cdot softmax(x)$ are more similar, compared to $x*softmax(x)$. The $x*softmax(x)$ may have some special attention patterns, as shown in layer 3 and layer 10. 

\paragraph{Attention scores can be negative, contrary to previous beliefs that attention scores must be positive.}
In prior work, the research community has attempted to replace softmax attention with ReLU attention or Sigmoid attention, operating under the assumption that attention scores should always be positive \cite{nair2010rectified,chen2020arelu,wortsman2023replacing,shen2023study} \cite{ramapuram2024theory} However, in this work, we successfully demonstrate that attention scores can indeed take on negative values. As shown in Appendix \ref{appendix: visualization attention}, we observe that transformers can still be effectively trained even when the attention scores contain negative elements and the sum of each row is not strictly equal to one.

\section{Conclusion}
We propose \methodFullName, a modification designed to improve gradient dynamics and enhance performance in transformers. To demonstrate the effectiveness of \methodShortName, we conduct extensive experiments, including analyses with various positional encodings, training lengths, and model sizes and different variants. Additionally, we evaluate \methodShortName on downstream tasks, where the variant $\frac{(x - \min(x_{\min}, 0))}{\max(0, x_{\max}) - \min(x_{\min}, 0)} \cdot softmax(x)$ consistently proves to be the most effective across diverse settings. This powerful adjustment significantly enhances transformer scalability and generalization, offering promising potential for a wide range of applications.

\section*{Limitations}

The proposed method needs to find the max and min values first for the normalization. Therefore, there may be additional costs.



\nocite{*}

\bibliography{custom}

\begin{thebibliography}{120}
\providecommand{\natexlab}[1]{#1}

\bibitem[{Ainslie et~al.(2023)Ainslie, Lei, de~Jong, Ontanon, Brahma, Zemlyanskiy, Uthus, Guo, Lee-Thorp, Tay et~al.}]{ainslie2023colt5}
Joshua Ainslie, Tao Lei, Michiel de~Jong, Santiago Ontanon, Siddhartha Brahma, Yury Zemlyanskiy, David Uthus, Mandy Guo, James Lee-Thorp, Yi~Tay, et~al. 2023.
\newblock Co{LT}5: Faster long-range transformers with conditional computation.
\newblock In \emph{The 2023 Conference on Empirical Methods in Natural Language Processing}.

\bibitem[{Alexey(2020)}]{alexey2020image}
Dosovitskiy Alexey. 2020.
\newblock An image is worth 16x16 words: Transformers for image recognition at scale.
\newblock \emph{arXiv preprint arXiv: 2010.11929}.

\bibitem[{Ba(2016)}]{ba2016layer}
Jimmy~Lei Ba. 2016.
\newblock Layer normalization.
\newblock \emph{arXiv preprint arXiv:1607.06450}.

\bibitem[{Bahdanau et~al.(2015)Bahdanau, Cho, and Bengio}]{bahdanau2014neural}
Dzmitry Bahdanau, Kyunghyun Cho, and Yoshua Bengio. 2015.
\newblock \href {https://arxiv.org/abs/1409.0473} {Neural machine translation by jointly learning to align and translate}.
\newblock In \emph{3rd International Conference on Learning Representations, ICLR 2015}.

\bibitem[{Beltagy et~al.(2020)Beltagy, Peters, and Cohan}]{beltagy2020longformer}
Iz~Beltagy, Matthew~E Peters, and Arman Cohan. 2020.
\newblock Longformer: The long-document transformer.
\newblock \emph{arXiv preprint arXiv:2004.05150}.

\bibitem[{Bengio et~al.(1994)Bengio, Simard, and Frasconi}]{bengio1994learning}
Yoshua Bengio, Patrice Simard, and Paolo Frasconi. 1994.
\newblock Learning long-term dependencies with gradient descent is difficult.
\newblock \emph{IEEE Transactions on Neural Networks}, 5(2):157--166.

\bibitem[{Bisk et~al.(2020)Bisk, Zellers, Le~Bras, Gao, and Choi}]{bisk2020piqa}
Yonatan Bisk, Rowan Zellers, Ronan Le~Bras, Jianfeng Gao, and Yejin Choi. 2020.
\newblock Piqa: Reasoning about physical commonsense in natural language.
\newblock In \emph{Proceedings of the Thirty-Fourth AAAI Conference on Artificial Intelligence}.

\bibitem[{Bjorck et~al.(2018)Bjorck, Gomes, Selman, and Weinberger}]{bjorck2018understanding}
Nils Bjorck, Carla~P Gomes, Bart Selman, and Kilian~Q Weinberger. 2018.
\newblock Understanding batch normalization.
\newblock \emph{Advances in neural information processing systems}, 31.

\bibitem[{Black et~al.(2022)Black, Biderman, Hallahan, Anthony, Gao, Golding, He, Leahy, McDonell, Phang et~al.}]{black2022gpt}
Sid Black, Stella Biderman, Eric Hallahan, Quentin Anthony, Leo Gao, Laurence Golding, Horace He, Connor Leahy, Kyle McDonell, Jason Phang, et~al. 2022.
\newblock Gpt-neox-20b: An open-source autoregressive language model.
\newblock \emph{arXiv preprint arXiv:2204.06745}.

\bibitem[{Blanchard et~al.(2019)Blanchard, Higham, and Higham}]{blanchard2019accurate}
Pierre Blanchard, Desmond~J Higham, and Nicholas~J Higham. 2019.
\newblock Accurate computation of the log-sum-exp and softmax functions.
\newblock \emph{arXiv preprint arXiv:1909.03469}.

\bibitem[{Blanchard et~al.(2021)Blanchard, Higham, and Higham}]{blanchard2021accurately}
Pierre Blanchard, Desmond~J Higham, and Nicholas~J Higham. 2021.
\newblock Accurately computing the log-sum-exp and softmax functions.
\newblock \emph{IMA Journal of Numerical Analysis}, 41(4):2311--2330.

\bibitem[{Boyd and Vandenberghe(2004)}]{boyd2004convex}
Stephen Boyd and Lieven Vandenberghe. 2004.
\newblock \emph{Convex optimization}.
\newblock Cambridge university press.

\bibitem[{Bradbury et~al.(2018)Bradbury, Frostig, Hawkins, Johnson, Leary, and Maclaurin}]{jax2018github}
James Bradbury, Roy Frostig, Peter Hawkins, Matthew~James Johnson, Chris Leary, and Dougal Maclaurin. 2018.
\newblock {JAX}: {A}utograd and {XLA}.
\newblock \url{https://github.com/google/jax}.
\newblock Accessed: 2024-09-25.

\bibitem[{Bridle(1990)}]{bridle1990probabilistic}
John~S Bridle. 1990.
\newblock Probabilistic interpretation of feedforward classification network outputs, with relationships to statistical pattern recognition.
\newblock In \emph{Neurocomputing: Algorithms, architectures and applications}, pages 227--236. Springer.

\bibitem[{Brown et~al.(2020)Brown, Mann, Ryder, Subbiah, Kaplan, Dhariwal, Neelakantan, Shyam, Sastry, Askell, Agarwal, Herbert-Voss, Krueger, Henighan, Child, Ramesh, Ziegler, Wu, Winter, Hesse, Chen, Sigler, Litwin, Gray, Chess, Clark, Berner, McCandlish, Radford, Sutskever, and Amodei}]{brown2020language}
Tom~B Brown, Benjamin Mann, Nick Ryder, Melanie Subbiah, Jared Kaplan, Prafulla Dhariwal, Arvind Neelakantan, Pranav Shyam, Girish Sastry, Amanda Askell, Sandhini Agarwal, Ariel Herbert-Voss, Gretchen Krueger, Tom Henighan, Rewon Child, Aditya Ramesh, Daniel~M Ziegler, Jeffrey Wu, Clemens Winter, Christopher Hesse, Mark Chen, Eric Sigler, Mateusz Litwin, Scott Gray, Benjamin Chess, Jack Clark, Christopher Berner, Sam McCandlish, Alec Radford, Ilya Sutskever, and Dario Amodei. 2020.
\newblock Language models are few-shot learners.
\newblock \emph{arXiv preprint arXiv:2005.14165}.

\bibitem[{Cettolo et~al.(2017)Cettolo, Federico, Bentivogli, Niehues, St{\"u}ker, Sudoh, Yoshino, and Federmann}]{cettolo-etal-2017-overview}
Mauro Cettolo, Marcello Federico, Luisa Bentivogli, Jan Niehues, Sebastian St{\"u}ker, Katsuhito Sudoh, Koichiro Yoshino, and Christian Federmann. 2017.
\newblock \href {https://aclanthology.org/2017.iwslt-1.1} {Overview of the {IWSLT} 2017 evaluation campaign}.
\newblock In \emph{Proceedings of the 14th International Conference on Spoken Language Translation}, pages 2--14, Tokyo, Japan. International Workshop on Spoken Language Translation.

\bibitem[{Chen et~al.(2020)Chen, Li, and Xu}]{chen2020arelu}
Dengsheng Chen, Jun Li, and Kai Xu. 2020.
\newblock Arelu: Attention-based rectified linear unit.
\newblock \emph{arXiv preprint arXiv:2006.13858}.

\bibitem[{Chen et~al.(2024{\natexlab{a}})Chen, Shi, Li, Gao, Ren, Chen, Jiang, Li, Liu, and Huang}]{chen2024sepllm}
Guoxuan Chen, Han Shi, Jiawei Li, Yihang Gao, Xiaozhe Ren, Yimeng Chen, Xin Jiang, Zhenguo Li, Weiyang Liu, and Chao Huang. 2024{\natexlab{a}}.
\newblock Sep{LLM}: Accelerate large language models by compressing one segment into one separator.
\newblock \emph{arXiv preprint arXiv:2412.12094}.

\bibitem[{Chen et~al.(2024{\natexlab{b}})Chen, YU, GE, Yao, Xie, Wang, Kwok, Luo, Lu, and Li}]{chen2024pixartalpha}
Junsong Chen, Jincheng YU, Chongjian GE, Lewei Yao, Enze Xie, Zhongdao Wang, James Kwok, Ping Luo, Huchuan Lu, and Zhenguo Li. 2024{\natexlab{b}}.
\newblock \href {https://openreview.net/forum?id=eAKmQPe3m1} {Pixart-\${\textbackslash}alpha\$: Fast training of diffusion transformer for photorealistic text-to-image synthesis}.
\newblock In \emph{The Twelfth International Conference on Learning Representations}.

\bibitem[{Chi et~al.(2022)Chi, Fan, Ramadge, and Rudnicky}]{chi2022kerple}
Ta-Chung Chi, Ting-Han Fan, Peter~J Ramadge, and Alexander Rudnicky. 2022.
\newblock Kerple: Kernelized relative positional embedding for length extrapolation.
\newblock \emph{Advances in Neural Information Processing Systems}, 35:8386--8399.

\bibitem[{Child et~al.(2019{\natexlab{a}})Child, Gray, Radford, and Sutskever}]{child2019generating}
Rewon Child, Scott Gray, Alec Radford, and Ilya Sutskever. 2019{\natexlab{a}}.
\newblock Generating long sequences with sparse transformers.
\newblock \emph{arXiv preprint arXiv:1904.10509}.

\bibitem[{Child et~al.(2019{\natexlab{b}})Child, Gray, Radford, and Sutskever}]{child2019sparse}
Rewon Child, Scott Gray, Alec Radford, and Ilya Sutskever. 2019{\natexlab{b}}.
\newblock \href {https://arxiv.org/abs/1904.10509} {Generating long sequences with sparse transformers}.
\newblock \emph{arXiv preprint arXiv:1904.10509}.

\bibitem[{Cho(2014)}]{cho2014learning}
Kyunghyun Cho. 2014.
\newblock Learning phrase representations using rnn encoder-decoder for statistical machine translation.
\newblock \emph{arXiv preprint arXiv:1406.1078}.

\bibitem[{Choromanski et~al.(2020)Choromanski, Likhosherstov, Dohan, Song, Gane, Sarlos, Hawkins, Davis, Mohiuddin, Kaiser et~al.}]{choromanski2020rethinking}
Krzysztof Choromanski, Valerii Likhosherstov, David Dohan, Xingyou Song, Andreea Gane, Tamas Sarlos, Peter Hawkins, Jared Davis, Afroz Mohiuddin, Lukasz Kaiser, et~al. 2020.
\newblock Rethinking attention with performers.
\newblock \emph{arXiv preprint arXiv:2009.14794}.

\bibitem[{Choromanski et~al.(2021)Choromanski, Likhosherstov, Dohan, Song, Gane, Sarlos, Hawkins, Davis, Mohiuddin, Kaiser et~al.}]{choromanski2021performer}
Krzysztof Choromanski, Valerii Likhosherstov, David Dohan, Xingyou Song, Andreea Gane, Tamas Sarlos, Peter Hawkins, Jared~Q Davis, Afroz Mohiuddin, {\L}ukasz Kaiser, et~al. 2021.
\newblock Rethinking attention with performers.
\newblock In \emph{International Conference on Learning Representations}.

\bibitem[{Clark et~al.(2019)Clark, Lee, Chang, Kwiatkowski, Collins, and Toutanova}]{clark2018boolq}
Christopher Clark, Kenton Lee, Ming-Wei Chang, Tom Kwiatkowski, Michael Collins, and Kristina Toutanova. 2019.
\newblock Boolq: Exploring the surprising difficulty of natural yes/no questions.
\newblock In \emph{Proceedings of the 2019 Conference of the North American Chapter of the Association for Computational Linguistics: Human Language Technologies}.

\bibitem[{Clark et~al.(2018{\natexlab{a}})Clark, Cowhey, Etzioni, Khot, Sabharwal, Schoenick, and Tafjord}]{clark2018think}
Peter Clark, Isaac Cowhey, Oren Etzioni, Tushar Khot, Ashish Sabharwal, Carissa Schoenick, and Oyvind Tafjord. 2018{\natexlab{a}}.
\newblock Think you have solved question answering? try arc, the ai2 reasoning challenge.
\newblock \emph{arXiv preprint arXiv:1803.05457}.

\bibitem[{Clark et~al.(2018{\natexlab{b}})Clark, Cowhey, Etzioni, Khot, Sabharwal, Schoenick, and Tafjord}]{clark2018arc}
Peter Clark, Isaac Cowhey, Oren Etzioni, Tushar Khot, Ashish Sabharwal, Caroline Schoenick, and Oyvind Tafjord. 2018{\natexlab{b}}.
\newblock Think you have solved question answering? try arc, the ai2 reasoning challenge.
\newblock \emph{arXiv preprint arXiv:1803.05457}.

\bibitem[{Cloud(2023)}]{google_tpuv5_2023}
Google Cloud. 2023.
\newblock Google cloud tpu v5e: Next-generation ai hardware for large-scale model training.
\newblock \url{https://cloud.google.com/blog/products/ai-machine-learning/introducing-tpu-v5e}.
\newblock Accessed: 2024-09-25.

\bibitem[{Dahl et~al.(2023)Dahl, Schneider, Nado, Agarwal, Sastry, Hennig, Medapati, Eschenhagen, Kasimbeg, Suo et~al.}]{dahl2023benchmarking}
George~E Dahl, Frank Schneider, Zachary Nado, Naman Agarwal, Chandramouli~Shama Sastry, Philipp Hennig, Sourabh Medapati, Runa Eschenhagen, Priya Kasimbeg, Daniel Suo, et~al. 2023.
\newblock Benchmarking neural network training algorithms.
\newblock \emph{arXiv preprint arXiv:2306.07179}.

\bibitem[{Dao(2023)}]{dao2023flashattention}
Tri Dao. 2023.
\newblock Flashattention-2: Faster attention with better parallelism and work partitioning.
\newblock \emph{arXiv preprint arXiv:2307.08691}.

\bibitem[{Dao et~al.(2022)Dao, Fu, Ermon, Rudra, and R{\'e}}]{dao2022flashattention}
Tri Dao, Dan Fu, Stefano Ermon, Atri Rudra, and Christopher R{\'e}. 2022.
\newblock Flashattention: Fast and memory-efficient exact attention with io-awareness.
\newblock \emph{Advances in Neural Information Processing Systems}, 35:16344--16359.

\bibitem[{Dao et~al.(2024)Dao, Fu, Wang et~al.}]{dao2024flashattention}
Tri Dao, Daniel Fu, Xinyang~G Wang, et~al. 2024.
\newblock \href {https://arxiv.org/abs/2401.14155} {Flashattention 2: Faster attention with better memory scheduling}.
\newblock \emph{arXiv preprint arXiv:2401.14155}.

\bibitem[{Dao and Gu(2024)}]{dao2024transformers}
Tri Dao and Albert Gu. 2024.
\newblock Transformers are ssms: Generalized models and efficient algorithms through structured state space duality.
\newblock \emph{arXiv preprint arXiv:2405.21060}.

\bibitem[{Deng et~al.(2009)Deng, Dong, Socher, Li, Li, and Fei-Fei}]{deng2009imagenet}
Jia Deng, Wei Dong, Richard Socher, Li-Jia Li, Kai Li, and Li~Fei-Fei. 2009.
\newblock Imagenet: A large-scale hierarchical image database.
\newblock In \emph{2009 IEEE conference on computer vision and pattern recognition}, pages 248--255. Ieee.

\bibitem[{Deng et~al.(2023)Deng, Song, and Zhou}]{deng2023superiority}
Yichuan Deng, Zhao Song, and Tianyi Zhou. 2023.
\newblock Superiority of softmax: Unveiling the performance edge over linear attention.
\newblock \emph{arXiv preprint arXiv:2310.11685}.

\bibitem[{Devlin et~al.(2018)Devlin, Chang, Lee, and Toutanova}]{devlin2018bert}
Jacob Devlin, Ming-Wei Chang, Kenton Lee, and Kristina Toutanova. 2018.
\newblock Bert: Pre-training of deep bidirectional transformers for language understanding.
\newblock \emph{arXiv preprint arXiv:1810.04805}.

\bibitem[{Dosovitskiy et~al.(2021)Dosovitskiy, Beyer, Kolesnikov, Weissenborn, Zhai, Unterthiner, Dehghani, Minderer, Heigold, Gelly et~al.}]{dosovitskiy2021vit}
Alexey Dosovitskiy, Lucas Beyer, Alexander Kolesnikov, Dirk Weissenborn, Xiaohua Zhai, Thomas Unterthiner, Mostafa Dehghani, Matthias Minderer, Georg Heigold, Sylvain Gelly, et~al. 2021.
\newblock An image is worth 16x16 words: Transformers for image recognition at scale.
\newblock \emph{arXiv preprint arXiv:2010.11929}.

\bibitem[{Dozat(2016)}]{dozat2016incorporating}
Timothy Dozat. 2016.
\newblock Incorporating nesterov momentum into adam.

\bibitem[{Dubey et~al.(2024)Dubey, Jauhri, Pandey, Kadian, Al-Dahle, Letman, Mathur, Schelten, Yang, Fan et~al.}]{dubey2024llama}
Abhimanyu Dubey, Abhinav Jauhri, Abhinav Pandey, Abhishek Kadian, Ahmad Al-Dahle, Aiesha Letman, Akhil Mathur, Alan Schelten, Amy Yang, Angela Fan, et~al. 2024.
\newblock The llama 3 herd of models.
\newblock \emph{arXiv preprint arXiv:2407.21783}.

\bibitem[{Duvvuri and Dhillon(2024)}]{duvvuri2024laser}
Sai~Surya Duvvuri and Inderjit~S Dhillon. 2024.
\newblock Laser: Attention with exponential transformation.
\newblock \emph{arXiv preprint arXiv:2411.03493}.

\bibitem[{Fountas et~al.(2024)Fountas, Benfeghoul, Oomerjee, Christopoulou, Lampouras, Bou-Ammar, and Wang}]{fountas2024human}
Zafeirios Fountas, Martin~A Benfeghoul, Adnan Oomerjee, Fenia Christopoulou, Gerasimos Lampouras, Haitham Bou-Ammar, and Jun Wang. 2024.
\newblock Human-like episodic memory for infinite context llms.
\newblock \emph{arXiv preprint arXiv:2407.09450}.

\bibitem[{Gao et~al.(2025)Gao, Zheng, Xie, Shi, Hu, Li, Ng, Li, and Liu}]{gaoalgoformer}
Yihang Gao, Chuanyang Zheng, Enze Xie, Han Shi, Tianyang Hu, Yu~Li, Michael Ng, Zhenguo Li, and Zhaoqiang Liu. 2025.
\newblock Algoformer: An efficient transformer framework with algorithmic structures.
\newblock \emph{Transactions on Machine Learning Research}.

\bibitem[{Gemini(2024)}]{gemini2024}
Team Gemini. 2024.
\newblock \href {https://arxiv.org/abs/2403.05530} {Gemini 1.5: Unlocking multimodal understanding across millions of tokens of context}.
\newblock \emph{arXiv preprint arXiv:2403.05530}.

\bibitem[{Geneva and Zabaras(2022)}]{geneva2022Transformers}
Nicholas Geneva and Nicholas Zabaras. 2022.
\newblock Transformers for modeling physical systems.
\newblock \emph{Neural Networks}, 146:272--289.

\bibitem[{Gilmer et~al.(2023)Gilmer, Dahl, Nado, Kasimbeg, and Medapati}]{init2winit2021github}
Justin~M. Gilmer, George~E. Dahl, Zachary Nado, Priya Kasimbeg, and Sourabh Medapati. 2023.
\newblock \href {http://github.com/google/init2winit} {{init2winit}: a jax codebase for initialization, optimization, and tuning research}.

\bibitem[{Glorot and Bengio(2010)}]{glorot2010understanding}
Xavier Glorot and Yoshua Bengio. 2010.
\newblock Understanding the difficulty of training deep feedforward neural networks.
\newblock In \emph{Proceedings of the thirteenth international conference on artificial intelligence and statistics}, pages 249--256. JMLR Workshop and Conference Proceedings.

\bibitem[{Google(2023)}]{pax2023github}
Research Google. 2023.
\newblock Pax: A {JAX}-based neural network training framework.
\newblock \url{https://github.com/google/paxml}.
\newblock Accessed: 2024-09-25.

\bibitem[{Graves and Graves(2012)}]{graves2012long}
Alex Graves and Alex Graves. 2012.
\newblock Long short-term memory.
\newblock \emph{Supervised sequence labelling with recurrent neural networks}, pages 37--45.

\bibitem[{Gu and Dao(2023)}]{gu2023mamba}
Albert Gu and Tri Dao. 2023.
\newblock Mamba: Linear-time sequence modeling with selective state spaces.
\newblock \emph{arXiv preprint arXiv:2312.00752}.

\bibitem[{Gulati et~al.(2020)Gulati, Qin, Chiu, Parmar, Zhang, Yu, Han, Wang, Zhang, Wu et~al.}]{gulati2020conformer}
Anmol Gulati, James Qin, Chung-Cheng Chiu, Niki Parmar, Yu~Zhang, Jiahui Yu, Wei Han, Shibo Wang, Zhengdong Zhang, Yonghui Wu, et~al. 2020.
\newblock Conformer: Convolution-augmented transformer for speech recognition.
\newblock In \emph{Proc. Interspeech}.

\bibitem[{Guo et~al.(2022)Guo, Ainslie, Uthus, Ontanon, Ni, Sung, and Yang}]{guo2021longt5}
Mandy Guo, Joshua Ainslie, David Uthus, Santiago Ontanon, Jianmo Ni, Yun-Hsuan Sung, and Yinfei Yang. 2022.
\newblock Long{T}5: Efficient text-to-text transformer for long sequences.
\newblock \emph{Findings of the Association for Computational Linguistics: NAACL}.

\bibitem[{Han et~al.(2024)Han, Pu, Xia, Han, Pan, Li, Lu, Song, and Huang}]{han2024bridging}
Dongchen Han, Yifan Pu, Zhuofan Xia, Yizeng Han, Xuran Pan, Xiu Li, Jiwen Lu, Shiji Song, and Gao Huang. 2024.
\newblock \href {https://openreview.net/forum?id=RSiGFzQapl} {Bridging the divide: Reconsidering softmax and linear attention}.
\newblock In \emph{The Thirty-eighth Annual Conference on Neural Information Processing Systems}.

\bibitem[{He et~al.(2016)He, Zhang, Ren, and Sun}]{he2016deep}
Kaiming He, Xiangyu Zhang, Shaoqing Ren, and Jian Sun. 2016.
\newblock Deep residual learning for image recognition.
\newblock In \emph{Proceedings of the IEEE conference on computer vision and pattern recognition}, pages 770--778.

\bibitem[{Hochreiter and Schmidhuber(1997)}]{hochreiter1997lstm}
Sepp Hochreiter and J{\"u}rgen Schmidhuber. 1997.
\newblock Long short-term memory.
\newblock \emph{Neural computation}, 9(8):1735--1780.

\bibitem[{Huang and Belongie(2017)}]{huang2017arbitrary}
Xun Huang and Serge Belongie. 2017.
\newblock Arbitrary style transfer in real-time with adaptive instance normalization.
\newblock In \emph{Proceedings of the IEEE international conference on computer vision}, pages 1501--1510.

\bibitem[{Ioffe(2015)}]{ioffe2015batch}
Sergey Ioffe. 2015.
\newblock Batch normalization: Accelerating deep network training by reducing internal covariate shift.
\newblock \emph{arXiv preprint arXiv:1502.03167}.

\bibitem[{Katharopoulos et~al.(2020)Katharopoulos, Vyas, Pappas, and Fleuret}]{katharopoulos2020transformers}
Angelos Katharopoulos, Apoorv Vyas, Nikolaos Pappas, and Fran{\c{c}}ois Fleuret. 2020.
\newblock Transformers are {RNNs}: Fast autoregressive transformers with linear attention.
\newblock In \emph{International conference on machine learning}, pages 5156--5165. PMLR.

\bibitem[{Khashabi et~al.(2018)Khashabi, Chaturvedi, Roth, Upadhyay, and Roth}]{khashabi2018multiRC}
Daniel Khashabi, Snigdha Chaturvedi, Michael Roth, Shyam Upadhyay, and Dan Roth. 2018.
\newblock Looking beyond the surface: A challenge set for reading comprehension over multiple sentences.
\newblock \emph{Proceedings of the 2018 Conference of the North American Chapter of the Association for Computational Linguistics: Human Language Technologies}, pages 252--262.

\bibitem[{Kocijan et~al.(2020)Kocijan, Chamorro-Perera, Sileo, Raiman, and Clark}]{kocijan2019winogrande}
Vid Kocijan, Elias Chamorro-Perera, Damien Sileo, Jonathan Raiman, and Peter Clark. 2020.
\newblock Winogrande: An adversarial winograd schema challenge at scale.
\newblock In \emph{Proceedings of the AAAI Conference on Artificial Intelligence}, volume~34, pages 8732--8740.

\bibitem[{Lai et~al.(2017)Lai, Xie, Liu, Yang, and Hovy}]{lai2017race}
Guokun Lai, Qizhe Xie, Hanxiao Liu, Yiming Yang, and Eduard Hovy. 2017.
\newblock Race: Large-scale reading comprehension dataset from examinations.
\newblock In \emph{Proceedings of the 2017 Conference on Empirical Methods in Natural Language Processing}, pages 785--794.

\bibitem[{LeCun et~al.(2002)LeCun, Bottou, Orr, and M{\"u}ller}]{lecun2002efficient}
Yann LeCun, L{\'e}on Bottou, Genevieve~B Orr, and Klaus-Robert M{\"u}ller. 2002.
\newblock Efficient backprop.
\newblock In \emph{Neural networks: Tricks of the trade}, pages 9--50. Springer.

\bibitem[{Levesque et~al.(2012)Levesque, Davis, and Morgenstern}]{levesque2012winograd}
Hector~J. Levesque, Ernest Davis, and Leora Morgenstern. 2012.
\newblock The winograd schema challenge.
\newblock In \emph{Proceedings of the Thirteenth International Conference on Principles of Knowledge Representation and Reasoning}, pages 552--561.

\bibitem[{Li et~al.(2023)Li, You, Guruganesh, Ainslie, Ontanon, Zaheer, Sanghai, Yang, Kumar, and Bhojanapalli}]{li2023functional}
Shanda Li, Chong You, Guru Guruganesh, Joshua Ainslie, Santiago Ontanon, Manzil Zaheer, Sumit Sanghai, Yiming Yang, Sanjiv Kumar, and Srinadh Bhojanapalli. 2023.
\newblock Functional interpolation for relative positions improves long context transformers.
\newblock \emph{arXiv preprint arXiv:2310.04418}.

\bibitem[{Lieber et~al.(2024)Lieber, Lenz, Bata, Cohen, Osin, Dalmedigos, Safahi, Meirom, Belinkov, Shalev-Shwartz et~al.}]{lieber2024jamba}
Opher Lieber, Barak Lenz, Hofit Bata, Gal Cohen, Jhonathan Osin, Itay Dalmedigos, Erez Safahi, Shaked Meirom, Yonatan Belinkov, Shai Shalev-Shwartz, et~al. 2024.
\newblock Jamba: A hybrid transformer-mamba language model.
\newblock \emph{arXiv preprint arXiv:2403.19887}.

\bibitem[{Lillicrap et~al.(2020)Lillicrap, Santoro, Marris, Akerman, and Hinton}]{lillicrap2020backpropagation}
Timothy~P Lillicrap, Adam Santoro, Luke Marris, Colin~J Akerman, and Geoffrey Hinton. 2020.
\newblock Backpropagation and the brain.
\newblock \emph{Nature Reviews Neuroscience}, 21(6):335--346.

\bibitem[{Liu et~al.(2023)Liu, Zaharia, and Abbeel}]{liu2023ring}
Hao Liu, Matei Zaharia, and Pieter Abbeel. 2023.
\newblock Ring attention with blockwise transformers for near-infinite context.
\newblock \emph{arXiv preprint arXiv:2310.01889}.

\bibitem[{Liu et~al.(2021{\natexlab{a}})Liu, Cui, Liu, Huang, Wang, and Zhang}]{liu2021logiqa}
Jian Liu, Leyang Cui, Hanmeng Liu, Dandan Huang, Yile Wang, and Yue Zhang. 2021{\natexlab{a}}.
\newblock Logiqa: a challenge dataset for machine reading comprehension with logical reasoning.
\newblock In \emph{Proceedings of the Twenty-Ninth International Conference on International Joint Conferences on Artificial Intelligence}, pages 3622--3628.

\bibitem[{Liu et~al.(2021{\natexlab{b}})Liu, Lin, Cao, Hu, Wei, Zhang, Lin, and Guo}]{liu2021swin}
Ze~Liu, Yutong Lin, Yue Cao, Han Hu, Yixuan Wei, Zheng Zhang, Stephen Lin, and Baining Guo. 2021{\natexlab{b}}.
\newblock Swin transformer: Hierarchical vision transformer using shifted windows.
\newblock In \emph{Proceedings of the IEEE/CVF international conference on computer vision}, pages 10012--10022.

\bibitem[{Liu et~al.(2021{\natexlab{c}})Liu, Huang, Huang, Li, and Zhao}]{liu2021finbert}
Zhuang Liu, Degen Huang, Kaiyu Huang, Zhuang Li, and Jun Zhao. 2021{\natexlab{c}}.
\newblock Finbert: A pre-trained financial language representation model for financial text mining.
\newblock In \emph{Proceedings of the Twenty-ninth International Conference on International Joint Conferences on Artificial Intelligence}, pages 4513--4519.

\bibitem[{Loshchilov and Hutter(2016)}]{loshchilov2016sgdr}
Ilya Loshchilov and Frank Hutter. 2016.
\newblock Sgdr: Stochastic gradient descent with warm restarts.
\newblock \emph{arXiv preprint arXiv:1608.03983}.

\bibitem[{Loshchilov and Hutter(2017)}]{loshchilov2017adamw}
Ilya Loshchilov and Frank Hutter. 2017.
\newblock Decoupled weight decay regularization.
\newblock \emph{arXiv preprint arXiv:1711.05101}.

\bibitem[{Merity et~al.(2017)Merity, Xiong, Bradbury, and Socher}]{merity17pointer}
Stephen Merity, Caiming Xiong, James Bradbury, and Richard Socher. 2017.
\newblock Pointer sentinel mixture models.
\newblock In \emph{International Conference on Learning Representations}.

\bibitem[{Meta-AI(2024)}]{llama3herd2024}
Team Meta-AI. 2024.
\newblock \href {https://arxiv.org/abs/2407.21783} {The llama 3 herd of models}.
\newblock \emph{arXiv preprint arXiv:2407.21783}.

\bibitem[{Mihaylov et~al.(2018{\natexlab{a}})Mihaylov, Clark, Khot, and Sabharwal}]{mihaylov2018openbookqa}
Todor Mihaylov, Peter Clark, Tushar Khot, and Ashish Sabharwal. 2018{\natexlab{a}}.
\newblock Can a suit of armor conduct electricity? a new dataset for open book question answering.
\newblock In \emph{Proceedings of the 2018 Conference on Empirical Methods in Natural Language Processing}, pages 2381--2391.

\bibitem[{Mihaylov et~al.(2018{\natexlab{b}})Mihaylov, Clark, Khot, and Sabharwal}]{mihaylov2018can}
Todor Mihaylov, Peter Clark, Tushar Khot, and Ashish Sabharwal. 2018{\natexlab{b}}.
\newblock Can a suit of armor conduct electricity? a new dataset for open book question answering.
\newblock In \emph{Proceedings of the 2018 Conference on Empirical Methods in Natural Language Processing}, pages 2381--2391.

\bibitem[{MLCommons()}]{mlcommons_bert_dataset}
MLCommons.
\newblock Bert dataset documentation.
\newblock \url{https://github.com/mlcommons/training/blob/master/language_model/tensorflow/bert/dataset.md}.
\newblock Accessed: 2024-10-16.

\bibitem[{Mostafazadeh et~al.(2016)Mostafazadeh, Chambers, He, Parikh, Batra, Vanderwende, Kohli, and Allen}]{mostafazadeh2016storycloze}
Nasrin Mostafazadeh, Nathanael Chambers, Xiaodong He, Devi Parikh, Dhruv Batra, Lucy Vanderwende, Pushmeet Kohli, and James Allen. 2016.
\newblock A corpus and cloze evaluation for deeper understanding of commonsense stories.
\newblock In \emph{Proceedings of the 2016 Conference of the North American Chapter of the Association for Computational Linguistics: Human Language Technologies}, pages 839--849.

\bibitem[{Nair and Hinton(2010)}]{nair2010rectified}
Vinod Nair and Geoffrey~E Hinton. 2010.
\newblock Rectified linear units improve restricted boltzmann machines.
\newblock In \emph{Proceedings of the 27th international conference on machine learning (ICML-10)}, pages 807--814.

\bibitem[{Nie et~al.(2019)Nie, Williams, Dinan, Bansal, Weston, and Kiela}]{nie2019anli}
Yixin Nie, Adina Williams, Emily Dinan, Mohit Bansal, Jason Weston, and Douwe Kiela. 2019.
\newblock Adversarial nli: A new benchmark for natural language understanding.
\newblock \emph{arXiv preprint arXiv:1910.14599}.

\bibitem[{Panayotov et~al.(2015)Panayotov, Chen, Povey, and Khudanpur}]{panayotov2015librispeech}
Vassil Panayotov, Guoguo Chen, Daniel Povey, and Sanjeev Khudanpur. 2015.
\newblock Librispeech: An asr corpus based on public domain audio books.
\newblock In \emph{2015 IEEE International Conference on Acoustics, Speech and Signal Processing (ICASSP)}, pages 5206--5210. IEEE.

\bibitem[{Paperno et~al.(2016)Paperno, Kruszewski, Lazaridou, Pham, Bernardi, Pezzelle, Baroni, Boleda, and Fern{\'a}ndez}]{paperno2016lambada}
Denis Paperno, Germ{\'a}n Kruszewski, Angeliki Lazaridou, Ngoc-Quan Pham, Raffaella Bernardi, Sandro Pezzelle, Marco Baroni, Gemma Boleda, and Raquel Fern{\'a}ndez. 2016.
\newblock The lambada dataset: Word prediction requiring a broad discourse context.
\newblock In \emph{Proceedings of the 54th Annual Meeting of the Association for Computational Linguistics (Volume 1: Long Papers)}, pages 1525--1534.

\bibitem[{Pascanu(2013)}]{pascanu2013difficulty}
R~Pascanu. 2013.
\newblock On the difficulty of training recurrent neural networks.
\newblock \emph{arXiv preprint arXiv:1211.5063}.

\bibitem[{Paszke et~al.(2019)Paszke, Gross, Massa, Lerer, Bradbury, Chanan, Killeen, Lin, Gimelshein, Antiga et~al.}]{paszke2019pytorch}
Adam Paszke, Sam Gross, Francisco Massa, Adam Lerer, James Bradbury, Gregory Chanan, Trevor Killeen, Zeming Lin, Natalia Gimelshein, Luca Antiga, et~al. 2019.
\newblock Pytorch: An imperative style, high-performance deep learning library.
\newblock \emph{Advances in Neural Information Processing Systems}, 32.

\bibitem[{Peebles and Xie(2023)}]{peebles2023scalable}
William Peebles and Saining Xie. 2023.
\newblock Scalable diffusion models with transformers.
\newblock In \emph{Proceedings of the IEEE/CVF International Conference on Computer Vision}, pages 4195--4205.

\bibitem[{Pilehvar and Camacho-Collados(2019)}]{pilehvar2019wic}
Mohammad~Taher Pilehvar and Jose Camacho-Collados. 2019.
\newblock Wic: The word-in-context dataset for evaluating context-sensitive meaning representations.
\newblock In \emph{Proceedings of the 2019 Conference of the North American Chapter of the Association for Computational Linguistics: Human Language Technologies, Volume 1 (Long and Short Papers)}, pages 1267--1273.

\bibitem[{Radford et~al.(2018)Radford, Narasimhan, Salimans, and Sutskever}]{radford2018improving}
Alec Radford, Karthik Narasimhan, Tim Salimans, and Ilya Sutskever. 2018.
\newblock Improving language understanding by generative pre-training.
\newblock OpenAI.

\bibitem[{Raffel et~al.(2020)Raffel, Shazeer, Roberts, Lee, Narang, Matena, Zhou, Li, and Liu}]{raffel2020exploring}
Colin Raffel, Noam Shazeer, Adam Roberts, Katherine Lee, Sharan Narang, Michael Matena, Yanqi Zhou, Wei Li, and Peter~J. Liu. 2020.
\newblock \href {http://jmlr.org/papers/v21/20-074.html} {Exploring the limits of transfer learning with a unified text-to-text transformer}.
\newblock In \emph{Journal of Machine Learning Research}, volume~21, pages 1--67.

\bibitem[{Ramapuram et~al.(2024)Ramapuram, Danieli, Dhekane, Weers, Busbridge, Ablin, Likhomanenko, Digani, Gu, Shidani et~al.}]{ramapuram2024theory}
Jason Ramapuram, Federico Danieli, Eeshan Dhekane, Floris Weers, Dan Busbridge, Pierre Ablin, Tatiana Likhomanenko, Jagrit Digani, Zijin Gu, Amitis Shidani, et~al. 2024.
\newblock Theory, analysis, and best practices for sigmoid self-attention.
\newblock \emph{arXiv preprint arXiv:2409.04431}.

\bibitem[{Roy et~al.(2021{\natexlab{a}})Roy, Saffar, Vaswani, and Grangier}]{roy2021efficient}
Aurko Roy, Mohammad Saffar, Ashish Vaswani, and David Grangier. 2021{\natexlab{a}}.
\newblock Efficient content-based sparse attention with routing transformers.
\newblock \emph{Transactions of the Association for Computational Linguistics}, 9:53--68.

\bibitem[{Roy et~al.(2021{\natexlab{b}})Roy, Saffar, Vaswani, and Grangier}]{roy2021routing}
Aurko Roy, Mohammad Saffar, Ashish Vaswani, and David Grangier. 2021{\natexlab{b}}.
\newblock \href {https://arxiv.org/abs/2003.05997} {Efficient routing transformers: Dynamic token interaction models for natural language processing}.
\newblock \emph{arXiv preprint arXiv:2003.05997}.

\bibitem[{Sakaguchi et~al.(2021)Sakaguchi, Bras, Bhagavatula, and Choi}]{sakaguchi2021winogrande}
Keisuke Sakaguchi, Ronan~Le Bras, Chandra Bhagavatula, and Yejin Choi. 2021.
\newblock Winogrande: An adversarial winograd schema challenge at scale.
\newblock \emph{Communications of the ACM}, 64(9):99--106.

\bibitem[{Salimans and Kingma(2016)}]{salimans2016weight}
Tim Salimans and Durk~P Kingma. 2016.
\newblock Weight normalization: A simple reparameterization to accelerate training of deep neural networks.
\newblock \emph{Advances in neural information processing systems}, 29.

\bibitem[{Shen et~al.(2023)Shen, Guo, Tan, Tang, Wang, and Bian}]{shen2023study}
Kai Shen, Junliang Guo, Xu~Tan, Siliang Tang, Rui Wang, and Jiang Bian. 2023.
\newblock A study on relu and softmax in transformer.
\newblock \emph{arXiv preprint arXiv:2302.06461}.

\bibitem[{Su et~al.(2024)Su, Ahmed, Lu, Pan, Bo, and Liu}]{su2024roformer}
Jianlin Su, Murtadha Ahmed, Yu~Lu, Shengfeng Pan, Wen Bo, and Yunfeng Liu. 2024.
\newblock Roformer: Enhanced transformer with rotary position embedding.
\newblock \emph{Neurocomputing}, 568:127063.

\bibitem[{Sun et~al.(2023)Sun, Zheng, Xie, Liu, Chu, Qiu, Xu, Ding, Li, Geng et~al.}]{sun2023survey}
Jiankai Sun, Chuanyang Zheng, Enze Xie, Zhengying Liu, Ruihang Chu, Jianing Qiu, Jiaqi Xu, Mingyu Ding, Hongyang Li, Mengzhe Geng, et~al. 2023.
\newblock A survey of reasoning with foundation models.
\newblock \emph{arXiv preprint arXiv:2312.11562}.

\bibitem[{Taylor et~al.(2022)Taylor, Kardas, Cucurull, Scialom, Hartshorn, Saravia, Poulton, Kerkez, and Stojnic}]{taylor2022galactica}
Ross Taylor, Marcin Kardas, Guillem Cucurull, Thomas Scialom, Anthony Hartshorn, Elvis Saravia, Andrew Poulton, Viktor Kerkez, and Robert Stojnic. 2022.
\newblock Galactica: A large language model for science.
\newblock \emph{arXiv preprint arXiv:2211.09085}.

\bibitem[{Touvron et~al.(2021)Touvron, Cord, Douze, Massa, Sablayrolles, and J{\'e}gou}]{touvron2021training}
Hugo Touvron, Matthieu Cord, Matthijs Douze, Francisco Massa, Alexandre Sablayrolles, and Herv{\'e} J{\'e}gou. 2021.
\newblock Training data-efficient image transformers \& distillation through attention.
\newblock In \emph{International conference on machine learning}, pages 10347--10357. PMLR.

\bibitem[{Vaswani et~al.(2017)Vaswani, Shazeer, Parmar, Uszkoreit, Jones, Gomez, Kaiser, and Polosukhin}]{vaswani2017attention}
Ashish Vaswani, Noam Shazeer, Niki Parmar, Jakob Uszkoreit, Llion Jones, Aidan~N Gomez, {\L}ukasz Kaiser, and Illia Polosukhin. 2017.
\newblock Attention is all you need.
\newblock \emph{Advances in Neural Information Processing Systems}, 30.

\bibitem[{Wang et~al.(2019)Wang, Pruksachatkun, Nangia, Singh, Michael, Hill, Levy, and Bowman}]{wang2019superglue}
Alex Wang, Yada Pruksachatkun, Nikita Nangia, Amanpreet Singh, Julian Michael, Felix Hill, Omer Levy, and Samuel~R Bowman. 2019.
\newblock Superglue: A stickier benchmark for general-purpose language understanding systems.
\newblock In \emph{Advances in Neural Information Processing Systems (NeurIPS)}, pages 3261--3275. Curran Associates, Inc.

\bibitem[{Wang et~al.(2020)Wang, Li, Khabsa, Fang, and Ma}]{wang2020linformer}
Sinong Wang, Belinda~Z Li, Madian Khabsa, Han Fang, and Hao Ma. 2020.
\newblock Linformer: Self-attention with linear complexity.
\newblock \emph{arXiv preprint arXiv:2006.04768}.

\bibitem[{Welbl et~al.(2017)Welbl, Liu, and Gardner}]{welbl2017crowdsourcing}
Johannes Welbl, Nelson~F Liu, and Matt Gardner. 2017.
\newblock Crowdsourcing multiple choice science questions.
\newblock In \emph{Proceedings of the 3rd Workshop on Noisy User-generated Text}, pages 94--106.

\bibitem[{Wortsman et~al.(2023)Wortsman, Lee, Gilmer, and Kornblith}]{wortsman2023replacing}
Mitchell Wortsman, Jaehoon Lee, Justin Gilmer, and Simon Kornblith. 2023.
\newblock Replacing softmax with relu in vision transformers.
\newblock \emph{arXiv preprint arXiv:2309.08586}.

\bibitem[{Wu et~al.(2023)Wu, Irsoy, Lu, Dabravolski, Dredze, Gehrmann, Kambadur, Rosenberg, and Mann}]{wu2023bloomberggpt}
Shijie Wu, Ozan Irsoy, Steven Lu, Vadim Dabravolski, Mark Dredze, Sebastian Gehrmann, Prabhanjan Kambadur, David Rosenberg, and Gideon Mann. 2023.
\newblock Bloomberggpt: A large language model for finance.
\newblock \emph{arXiv preprint arXiv:2303.17564}.

\bibitem[{Xiao et~al.(2024{\natexlab{a}})Xiao, Zhang, Han, Xiao, Lin, Zhang, Liu, Han, and Sun}]{xiao2024infllm}
Chaojun Xiao, Pengle Zhang, Xu~Han, Guangxuan Xiao, Yankai Lin, Zhengyan Zhang, Zhiyuan Liu, Song Han, and Maosong Sun. 2024{\natexlab{a}}.
\newblock {InfLLM}: Unveiling the intrinsic capacity of {LLMs} for understanding extremely long sequences with training-free memory.
\newblock \emph{arXiv preprint arXiv:2402.04617}.

\bibitem[{Xiao et~al.(2024{\natexlab{b}})Xiao, Tian, Chen, Han, and Lewis}]{xiao2024efficient}
Guangxuan Xiao, Yuandong Tian, Beidi Chen, Song Han, and Mike Lewis. 2024{\natexlab{b}}.
\newblock \href {https://openreview.net/forum?id=NG7sS51zVF} {Efficient streaming language models with attention sinks}.
\newblock In \emph{The Twelfth International Conference on Learning Representations}.

\bibitem[{Xiong et~al.(2025)Xiong, Shen, Zheng, Wan, Zhao, Yang, Ye, Yang, Kong, and Wong}]{xiong2025parallelcomp}
Jing Xiong, Jianghan Shen, Chuanyang Zheng, Zhongwei Wan, Chenyang Zhao, Chiwun Yang, Fanghua Ye, Hongxia Yang, Lingpeng Kong, and Ngai Wong. 2025.
\newblock Parallelcomp: Parallel long-context compressor for length extrapolation.
\newblock \emph{arXiv preprint arXiv:2502.14317}.

\bibitem[{Xiong et~al.(2020)Xiong, Yang, He, Zheng, Zheng, Lan, Wang, and Liu}]{xiong2020layer}
Ruibin Xiong, Yingquan Yang, Di~He, Kai Zheng, Shuxin Zheng, Yaliang Lan, Jingdong Wang, and Tie-Yan Liu. 2020.
\newblock On layer normalization in the transformer architecture.
\newblock In \emph{International Conference on Machine Learning}, pages 10524--10533. PMLR.

\bibitem[{Xu et~al.(2019)Xu, Sun, Zhang, Zhao, and Lin}]{xu2019understanding}
Jingjing Xu, Xu~Sun, Zhiyuan Zhang, Guangxiang Zhao, and Junyang Lin. 2019.
\newblock Understanding and improving layer normalization.
\newblock \emph{Advances in neural information processing systems}, 32.

\bibitem[{You et~al.(2019)You, Li, Reddi, Hseu, Kumar, Bhojanapalli, Song, Demmel, and Hsieh}]{you2019lamb}
Yang You, Jing Li, Sashank Reddi, Jason Hseu, Sanjiv Kumar, Srinadh Bhojanapalli, Xiaodan Song, James Demmel, and Cho-Jui Hsieh. 2019.
\newblock Large batch optimization for deep learning: Training bert in 76 minutes.
\newblock \emph{arXiv preprint arXiv:1904.00962}.

\bibitem[{Zellers et~al.(2019)Zellers, Holtzman, Bisk, Farhadi, and Choi}]{zellers2019hellaswag}
Rowan Zellers, Ari Holtzman, Yonatan Bisk, Ali Farhadi, and Yejin Choi. 2019.
\newblock Hellaswag: Can a machine really finish your sentence?
\newblock In \emph{Proceedings of the 57th Annual Meeting of the Association for Computational Linguistics}, pages 4791--4800.

\bibitem[{Zhang et~al.(2020)Zhang, Zhao, Saleh, and Liu}]{zhang2020pegasus}
Jingqing Zhang, Yao Zhao, Mohammad Saleh, and Peter Liu. 2020.
\newblock {PEGASUS}: Pre-training with extracted gap-sentences for abstractive summarization.
\newblock In \emph{International Conference on Machine Learning}, pages 11328--11339. PMLR.

\bibitem[{Zhang et~al.(2019)Zhang, He, Sra, and Jadbabaie}]{zhang2019gradient}
Jingzhao Zhang, Tianxing He, Suvrit Sra, and Ali Jadbabaie. 2019.
\newblock Why gradient clipping accelerates training: A theoretical justification for adaptivity.
\newblock \emph{arXiv preprint arXiv:1905.11881}.

\bibitem[{Zhang et~al.(2018)Zhang, Liu, Liu, Wang, Liu, Gao, Xu, Xu, Sun, Cui et~al.}]{zhang2018record}
Shuailong Zhang, Huanbo Liu, Shuyan Liu, Yuwei Wang, Jiawei Liu, Zhiyu Gao, Wei Xu, Yiming Xu, Xin Sun, Lei Cui, et~al. 2018.
\newblock Record: Bridging the gap between human and machine commonsense reading comprehension.
\newblock \emph{arXiv preprint arXiv:1810.12885}.

\bibitem[{Zhang et~al.(2015)Zhang, Zhao, and LeCun}]{zhang2015character}
Xiang Zhang, Junbo Zhao, and Yann LeCun. 2015.
\newblock Character-level convolutional networks for text classification.
\newblock \emph{Advances in neural information processing systems}, 28.

\bibitem[{Zheng et~al.(2024{\natexlab{a}})Zheng, Gao, Shi, Huang, Li, Xiong, Ren, Ng, Jiang, Li et~al.}]{zheng2024dape}
Chuanyang Zheng, Yihang Gao, Han Shi, Minbin Huang, Jingyao Li, Jing Xiong, Xiaozhe Ren, Michael Ng, Xin Jiang, Zhenguo Li, et~al. 2024{\natexlab{a}}.
\newblock Dape: Data-adaptive positional encoding for length extrapolation.
\newblock In \emph{The Thirty-eighth Annual Conference on Neural Information Processing Systems}.

\bibitem[{Zheng et~al.(2024{\natexlab{b}})Zheng, Gao, Shi, Xiong, Sun, Li, Huang, Ren, Ng, Jiang et~al.}]{zheng2024dapev2}
Chuanyang Zheng, Yihang Gao, Han Shi, Jing Xiong, Jiankai Sun, Jingyao Li, Minbin Huang, Xiaozhe Ren, Michael Ng, Xin Jiang, et~al. 2024{\natexlab{b}}.
\newblock Dape v2: Process attention score as feature map for length extrapolation.
\newblock \emph{arXiv preprint arXiv:2410.04798}.

\bibitem[{Zheng et~al.(2023)Zheng, Liu, Xie, Li, and Li}]{zheng2023progressive}
Chuanyang Zheng, Zhengying Liu, Enze Xie, Zhenguo Li, and Yu~Li. 2023.
\newblock Progressive-hint prompting improves reasoning in large language models.
\newblock \emph{arXiv preprint arXiv:2304.09797}.

\bibitem[{Zhou et~al.(2024)Zhou, Staats, Li, Szegedy, Weinberger, and Wu}]{zhou2024dont}
Jin~Peng Zhou, Charles~E Staats, Wenda Li, Christian Szegedy, Kilian~Q Weinberger, and Yuhuai Wu. 2024.
\newblock \href {https://openreview.net/forum?id=V5tdi14ple} {Don't trust: Verify -- grounding {LLM} quantitative reasoning with autoformalization}.
\newblock In \emph{The Twelfth International Conference on Learning Representations}.

\bibitem[{Zhu et~al.(2024)Zhu, Duan, Chen, Liu, Li, Feng, Lv, Cao, Chuanfu, Zhang et~al.}]{zhu2024near}
Qianchao Zhu, Jiangfei Duan, Chang Chen, Siran Liu, Xiuhong Li, Guanyu Feng, Xin Lv, Huanqi Cao, Xiao Chuanfu, Xingcheng Zhang, et~al. 2024.
\newblock Near-lossless acceleration of long context llm inference with adaptive structured sparse attention.
\newblock \emph{arXiv preprint arXiv:2406.15486}.

\end{thebibliography}

\clearpage
\newpage
\appendix

\section{Model Configuration} 
\label{appendix: experiment setting} 
\paragraph{Pretrain Setting.} All experiments are conducted on 8 GPUs. The 125M and 350M model configurations are the following.

\begin{table}[!ht]
    \centering
    \setlength{\tabcolsep}{2pt}
    \label{model configuration}
    \caption{\textbf{Model Configurations.}}
    \begin{tabular}{c c c c c}
    \toprule
    & & \textbf{125M} & & \textbf{350M} \\ \midrule
    Training sequence length & & $512$ & & $512$\\
    Batch size & & 2 $\times$ 8  & & 2 $\times$ 8 \\
    Number of iterations & & $50$k & & $50$k \\
    Dropout prob. & & $0.0$ & & $0.0$ \\
    Attention dropout prob. & & $0.0$ & & $0.0$ \\
    Attention head && 12 && 16  \\
    Feature dimension && 768 && 1024\\
    Layer number && 12 && 24 \\
    Optimizer & & Adam & & Adam\\
    Optimizer parameter betas & & [0.9, 0.95] && [0.9, 0.95] \\
    Learning rate & & $6\mathrm{e}-4$  & & $3\mathrm{e}-4$ \\
    Precision & & float132 & & float32 \\ 
    \bottomrule
    \end{tabular}
    \label{tab:model_configs}
\end{table}

\paragraph{Experiment Setting for Classification and Translation tasks.}
For the sequence classification tasks presented in Table \ref{table: classification_task}, the feature dimension is set to 128, with 2 attention heads and 6 layers. The datasets are AGNews, DBPedia, Yelp-Review, YahooNews, AmazonNews \cite{zhang2015character}.
In contrast, for the machine translation tasks shown in Table \ref{table: machine_translation}, the feature dimension is increased to 512, with 8 attention heads and 12 layers. The dataset comes from IWSLT2017 datasets \cite{cettolo-etal-2017-overview}.


\section{Error Bar}
\begin{table}[htb]
\caption{The perplexity on Books3 dataset with three random seeds.}
\centering
\resizebox{0.49\textwidth}{!}{
\begin{tabular}{cccc}
\toprule
\textbf{Data}&\textbf{\methodShortName} & Mean & Std  \\ \midrule
Kerple &\ding{53} &38.21 &0.3873 \\  
Kerple& \ding{51} &37.71 &0.3826\\ 
FIRE &\ding{53} &38.00  &0.2211 \\  
FIRE& \ding{51} &37.24 &0.2786\\ 
RoPE &\ding{53} &38.03 &0.2165 \\  
RoPE& \ding{51} &37.48 &0.3287\\ 
DAPEV2-Kerple &\ding{53} &35.92 &0.3821\\  
DAPEV2-Kerple &\ding{51} &35.58 &0.4037 \\  
\bottomrule
\end{tabular}
}
\label{table: sa-softmax}
\end{table}

\section{Analyze the Training Loss and Gradient}
\label{appendix: loss and gradient}
\begin{figure}[!htbp]
\setlength{\abovecaptionskip}{0.1cm}
\centering
\includegraphics[width=0.4\textwidth]{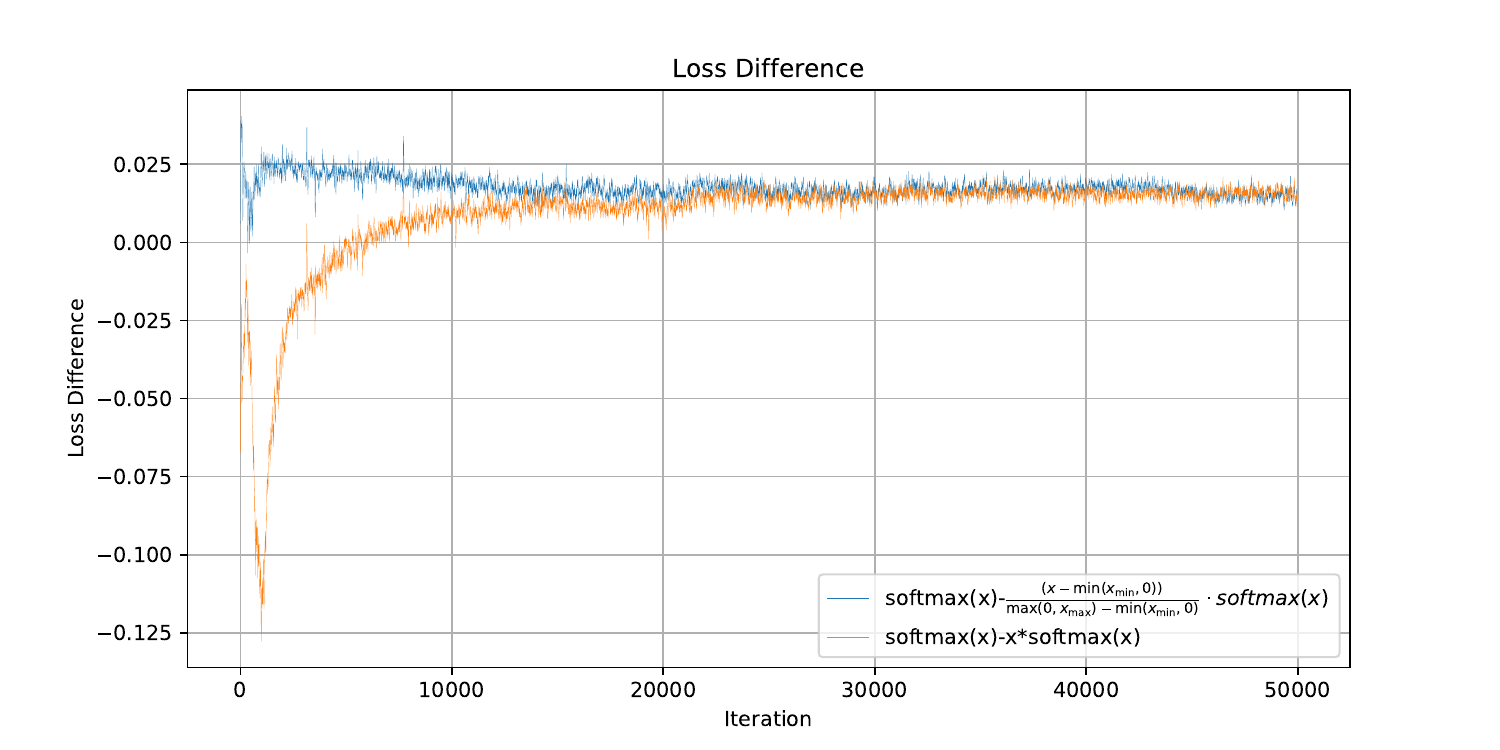}
\hspace{0in}
\includegraphics[width=0.4\textwidth]{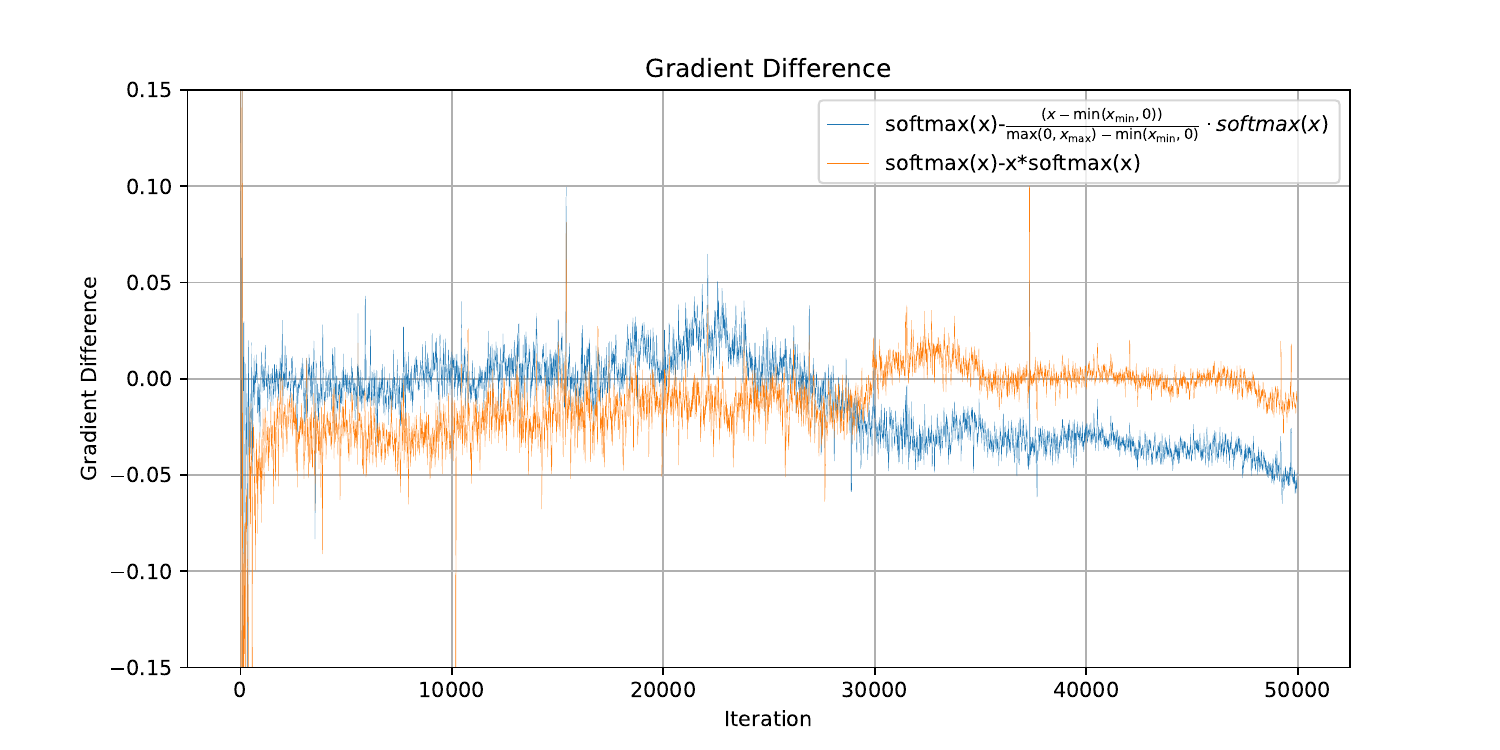}
\caption{
\small
The loss difference and gradient difference between our methods and baseline. 
}
\label{fig: dif_loss}
\end{figure}
\paragraph{Optimizer Gradient.} As shown in Figure \ref{fig: dif_loss}, comparing the gradients of $softmax(x)$ and $x \cdot softmax(x)$, the latter shows larger gradients initially due to the multiplicative factor of $x$. A detailed analysis in the methodology section confirms this behavior. In contrast, the normalized variant, $\frac{(x - \min(x_{\min}, 0))}{\max(0, x_{\max}) - \min(x_{\min}, 0)} \cdot softmax(x)$, produces gradients similar to or smaller than $softmax(x)$ early on but can grow larger later in training. This is due to normalization, which stabilizes updates but reduces gradient magnitude in the early stages.

\paragraph{Training Loss Across Steps.} For DAPEV2-Kerple, $x \cdot softmax(x)$ and $\frac{(x - \min(x_{\min}, 0))}{\max(0, x_{\max}) - \min(x_{\min}, 0)} \cdot softmax(x)$ can both achieve lower loss than baseline $softmax(x)$. However, $\frac{(x - \min(x_{\min}, 0))}{\max(0, x_{\max}) - \min(x_{\min}, 0)} \cdot softmax(x)$ is better than baseline $softmax(x)$ through the whole training steps, while the $x \cdot softmax(x)$ achieves better performance than bseline at late training step. This may caused by that the $\frac{(x - \min(x_{\min}, 0))}{\max(0, x_{\max}) - \min(x_{\min}, 0)} \cdot softmax(x)$ is a normalized version so that easier for optimizer. 

\section{Risk}
This work focuses on utilizing self-adjust softmax to improve the transformer architecture. This is no specific risk. Also, we use AI assistants for writing.
\onecolumn
\section{Visualization of Attention Score}
\label{appendix: visualization attention}
\begin{figure}[!htbp]
\setlength{\abovecaptionskip}{0.1cm}
\centering
\includegraphics[width=0.32\textwidth]{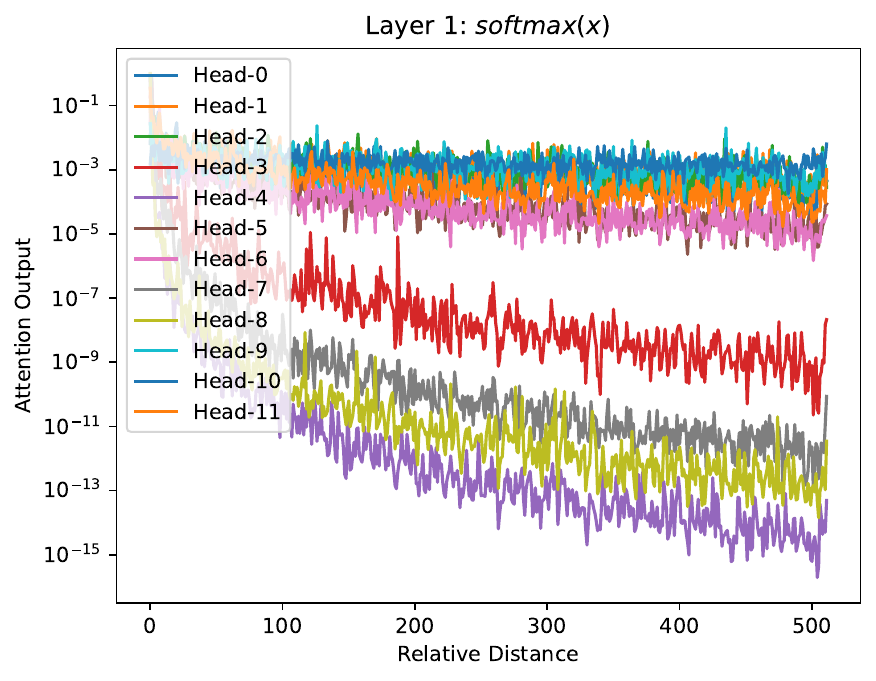}
\hspace{0in}
\includegraphics[width=0.32\textwidth]{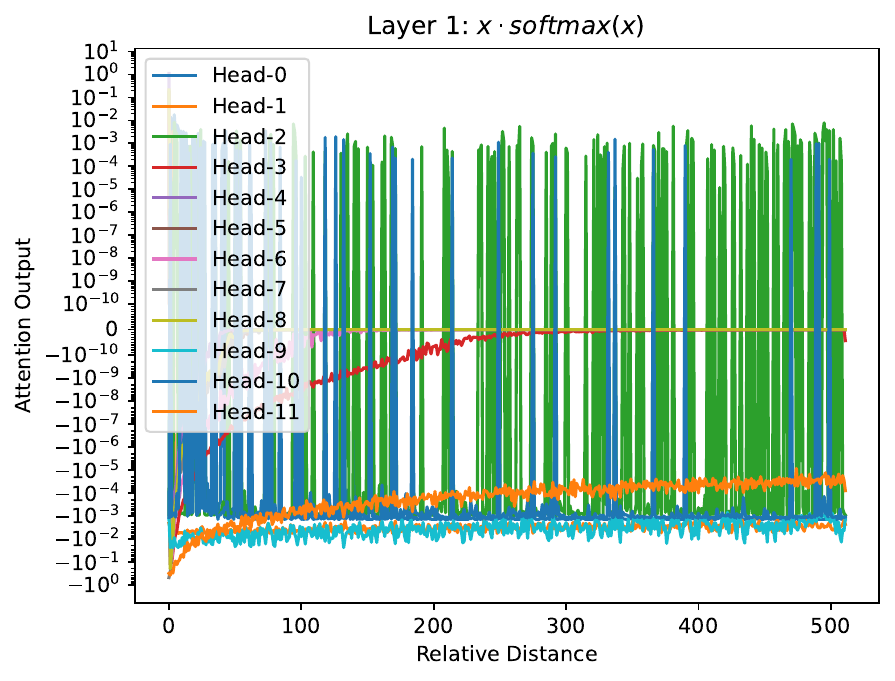}
\hspace{0in}
\includegraphics[width=0.32\textwidth]{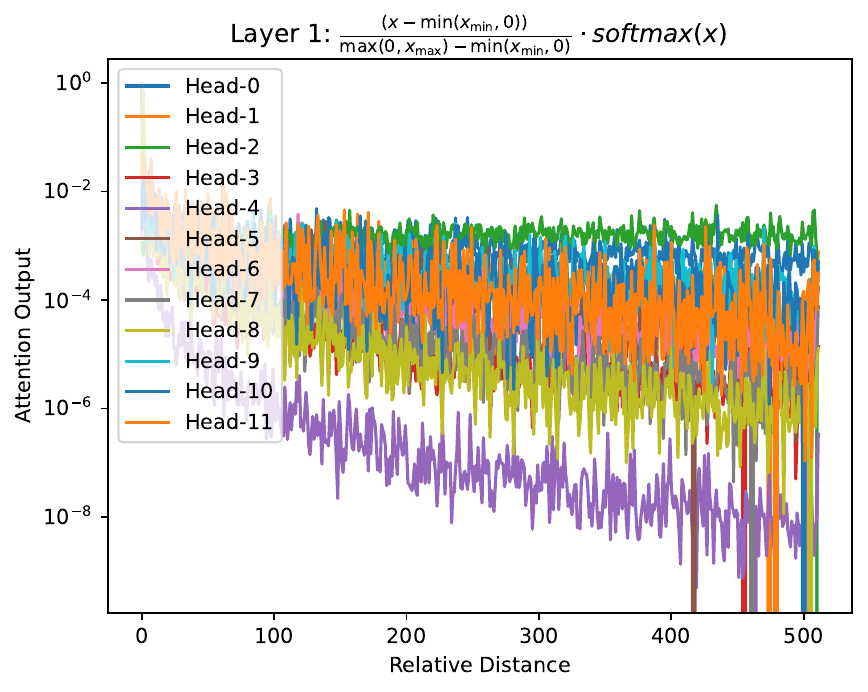}
\hspace{0in}

\includegraphics[width=0.32\textwidth]{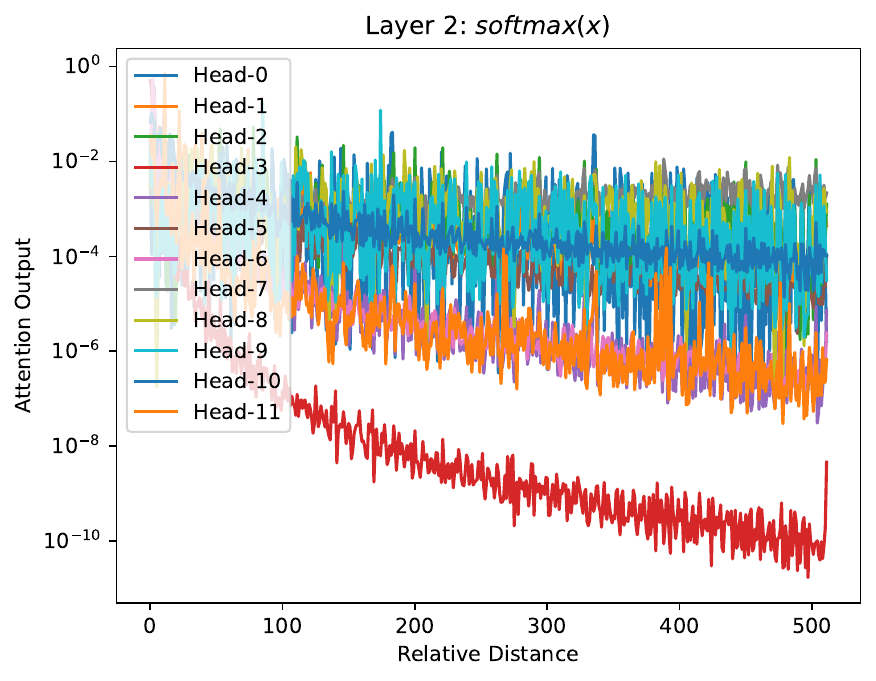}
\hspace{0in}
\includegraphics[width=0.32\textwidth]{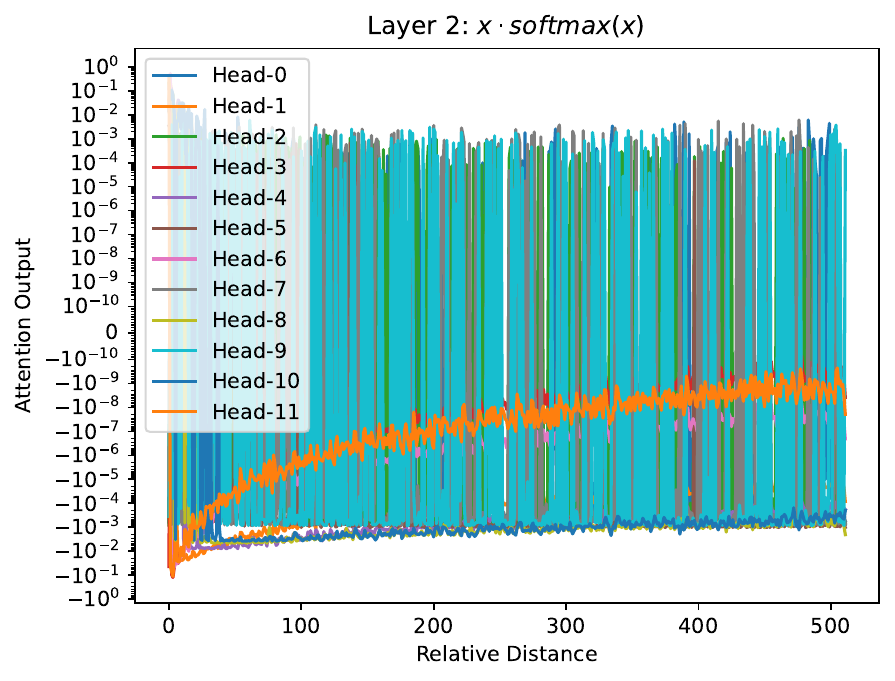}
\hspace{0in}
\includegraphics[width=0.32\textwidth]{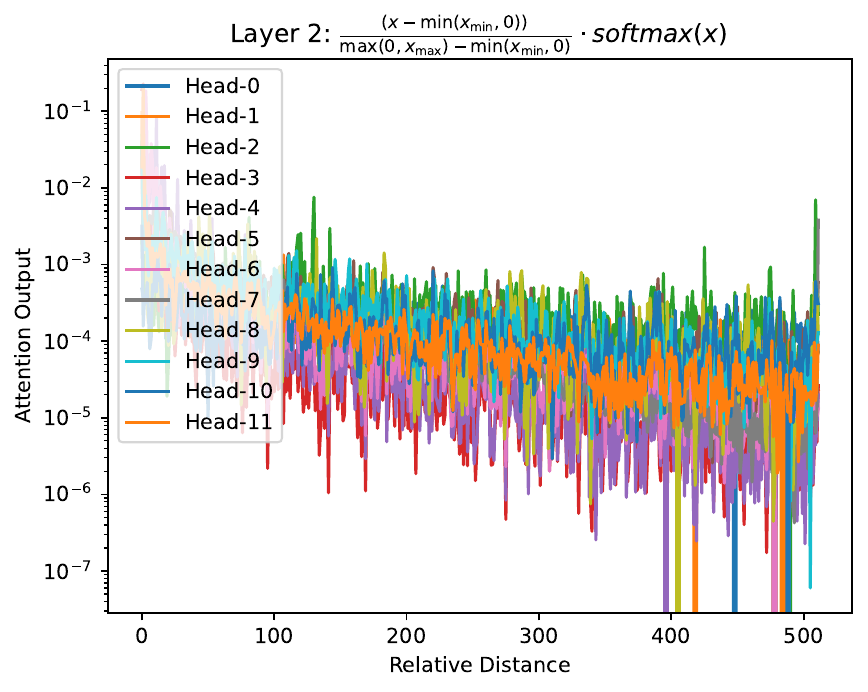}
\hspace{0in}

\includegraphics[width=0.32\textwidth]{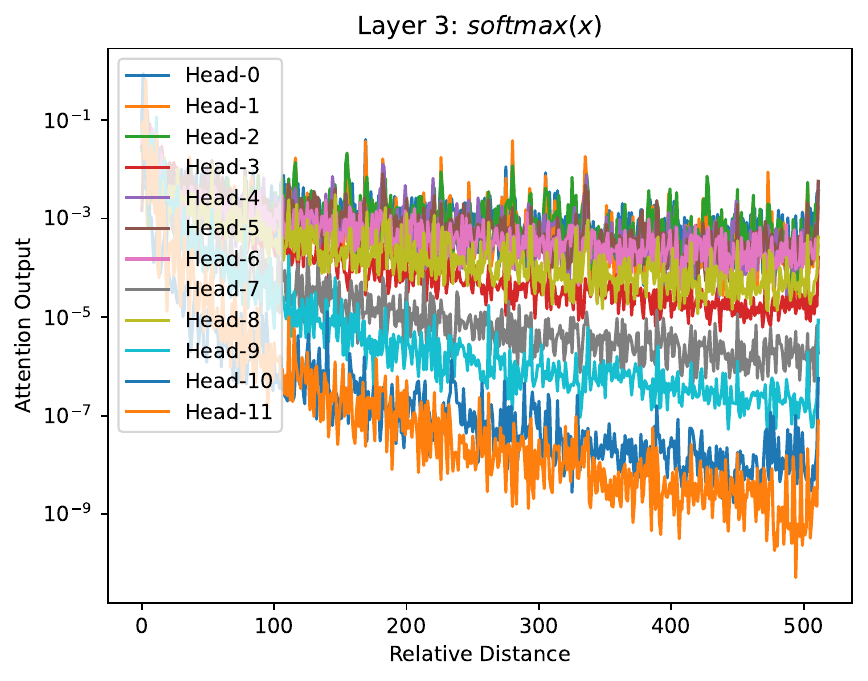}
\hspace{0in}
\includegraphics[width=0.32\textwidth]{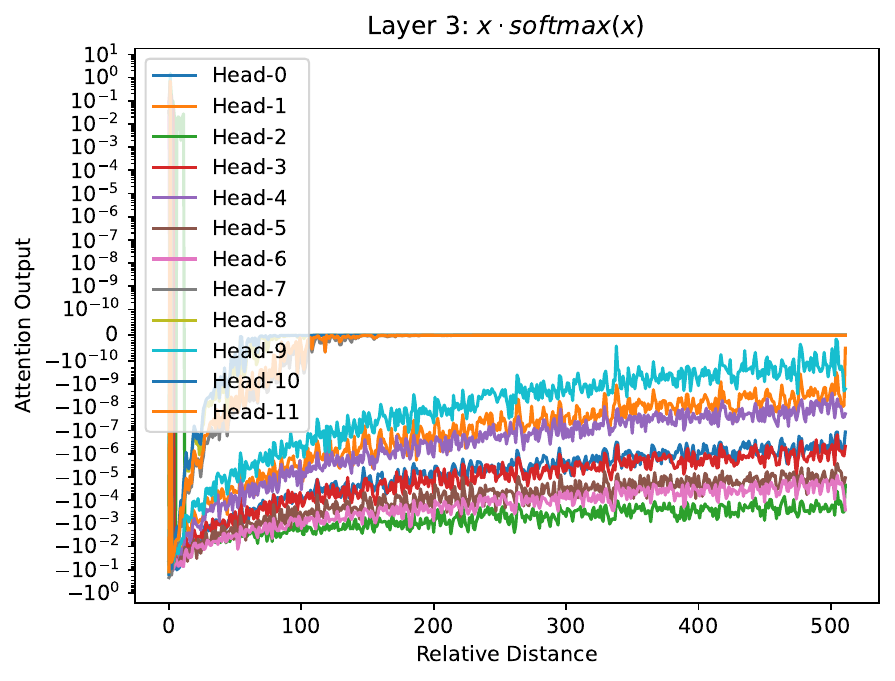}
\hspace{0in}
\includegraphics[width=0.32\textwidth]{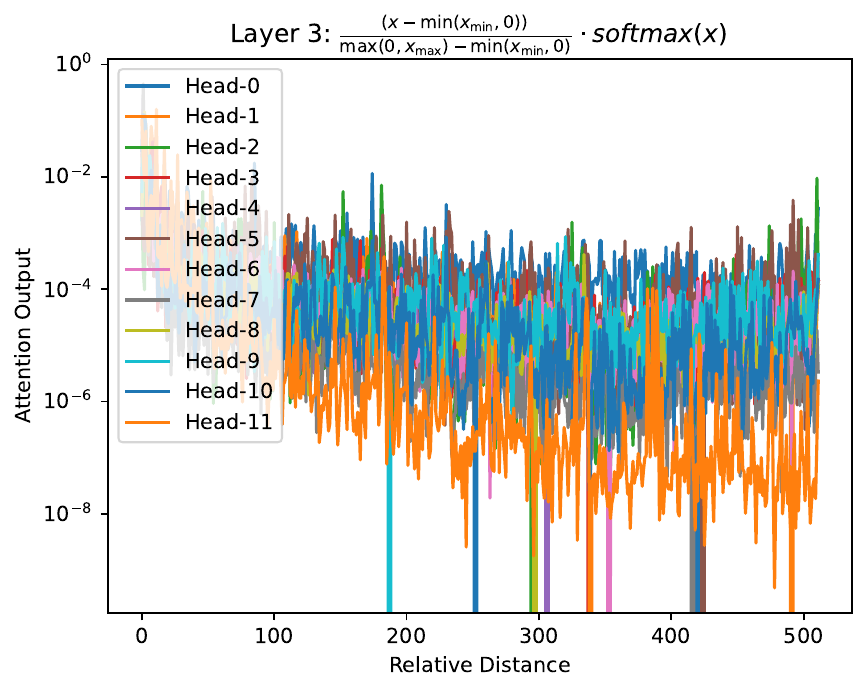}
\hspace{0in}

\includegraphics[width=0.32\textwidth]{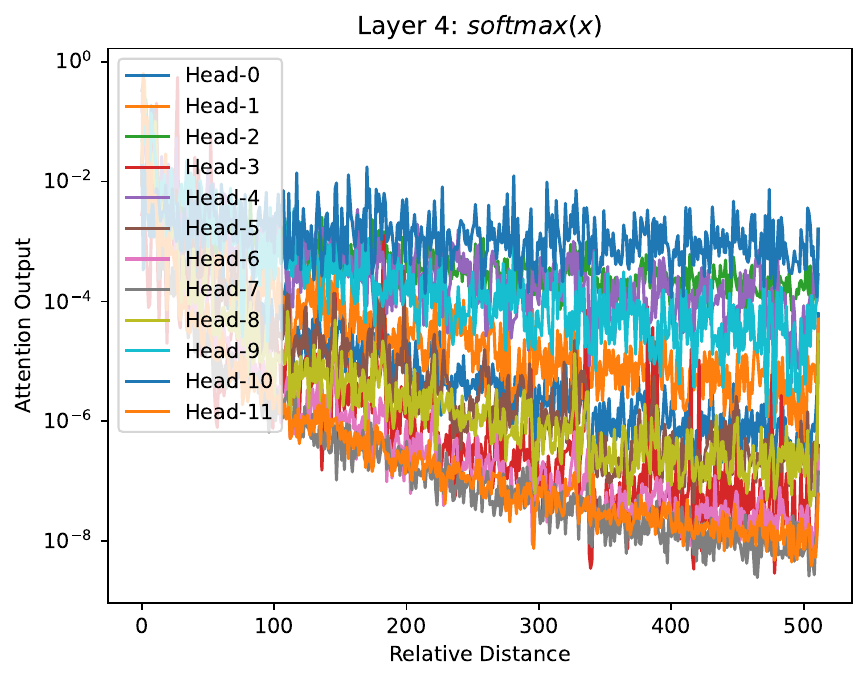}
\hspace{0in}
\includegraphics[width=0.32\textwidth]{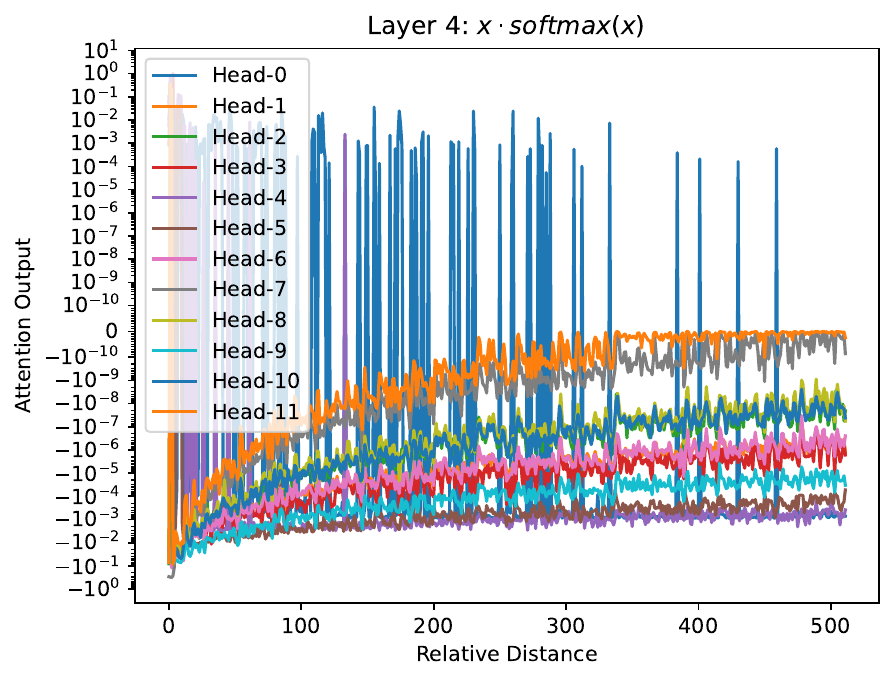}
\hspace{0in}
\includegraphics[width=0.32\textwidth]{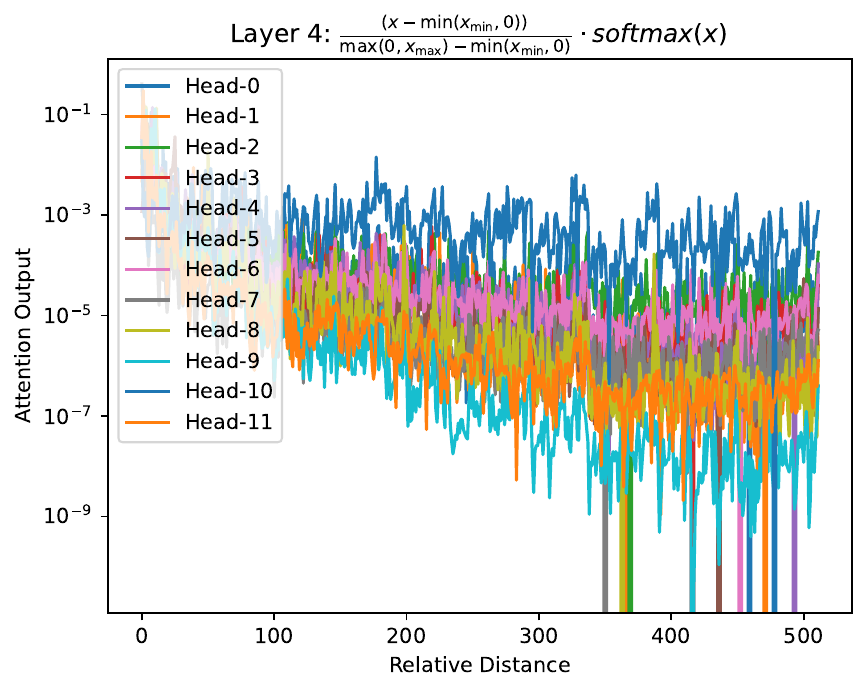}
\hspace{0in}

\setlength{\abovecaptionskip}{0.1cm}
\centering
\includegraphics[width=0.32\textwidth]{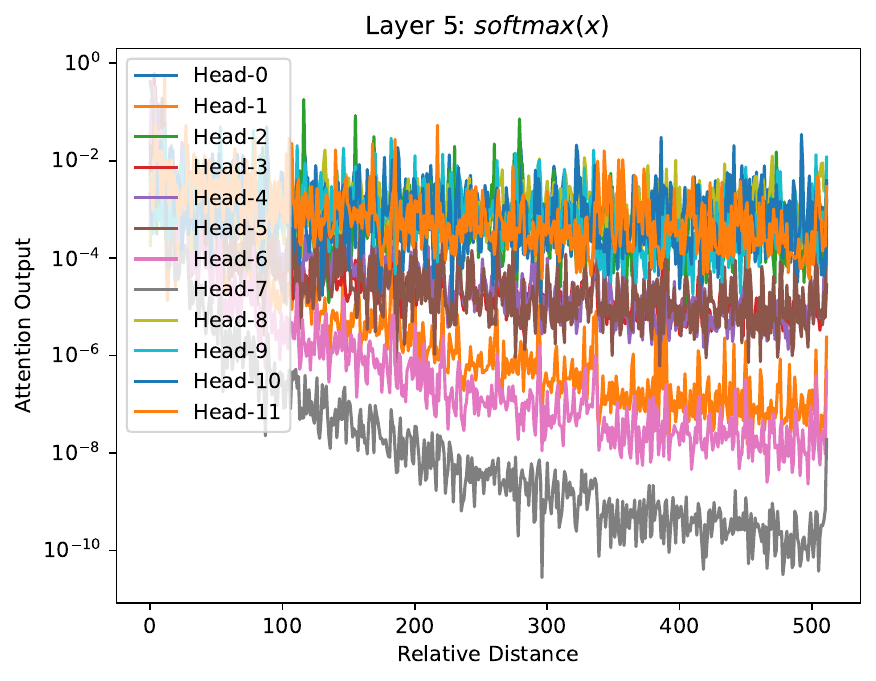}
\hspace{0in}
\includegraphics[width=0.32\textwidth]{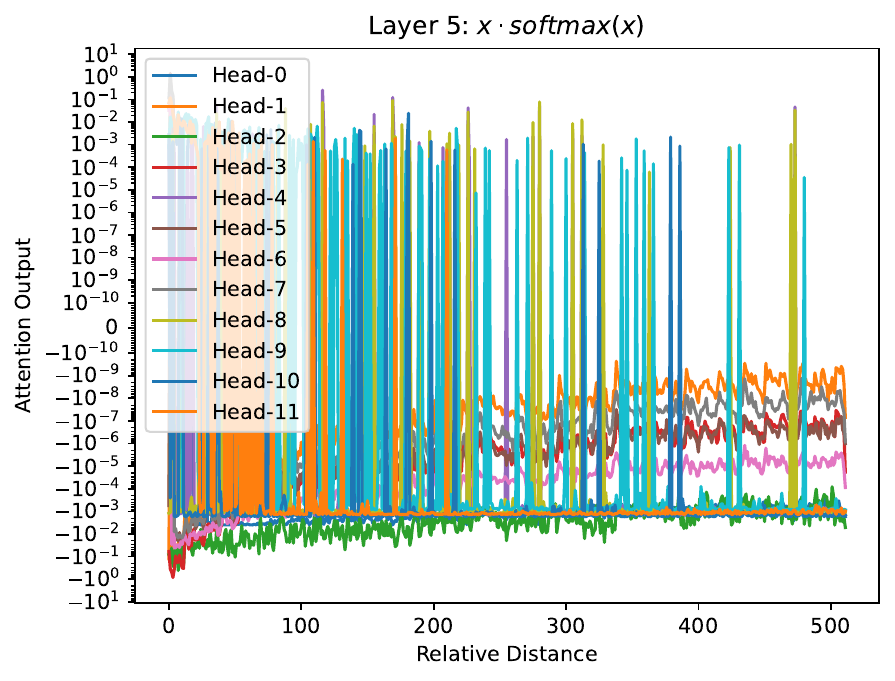}
\hspace{0in}
\includegraphics[width=0.32\textwidth]{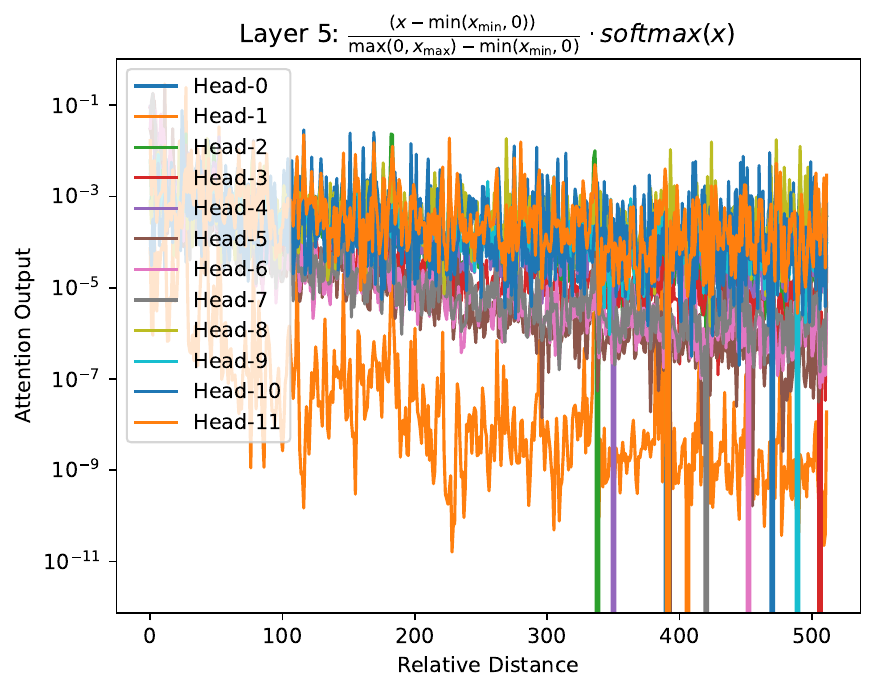}
\hspace{0in}

\caption{
\small
\textbf{The visualization of attention output, from left to right: 1) $softmax(x)$; 2) $x*softmax(x)$; 3) $\frac{(x - min(x_{\min},0))}{max(0,x_{max})-min(x_{min},0)} \cdot softmax(x)$.
}
}

\end{figure}

\begin{figure}[!htbp]




\includegraphics[width=0.32\textwidth]{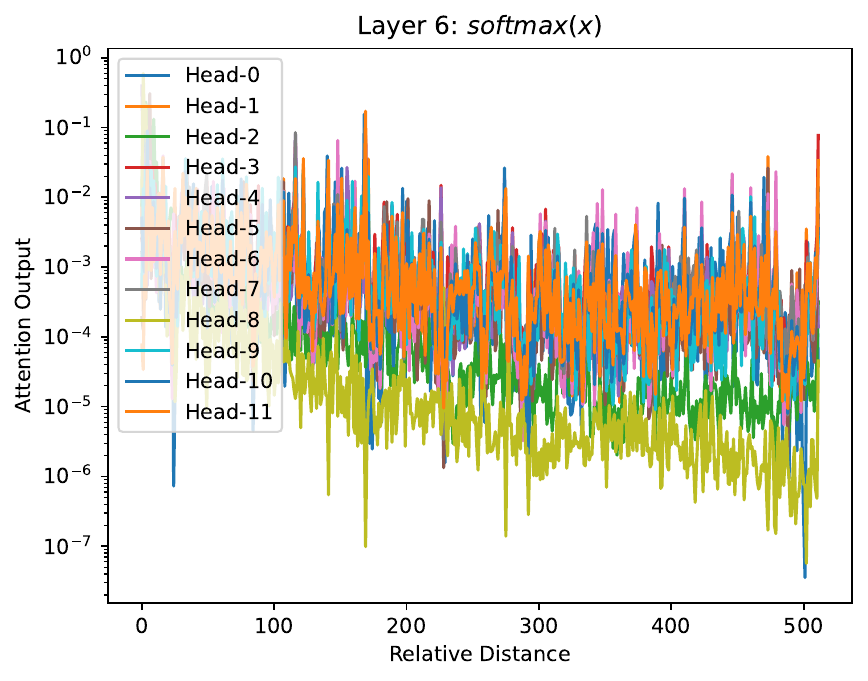}
\hspace{0in}
\includegraphics[width=0.32\textwidth]{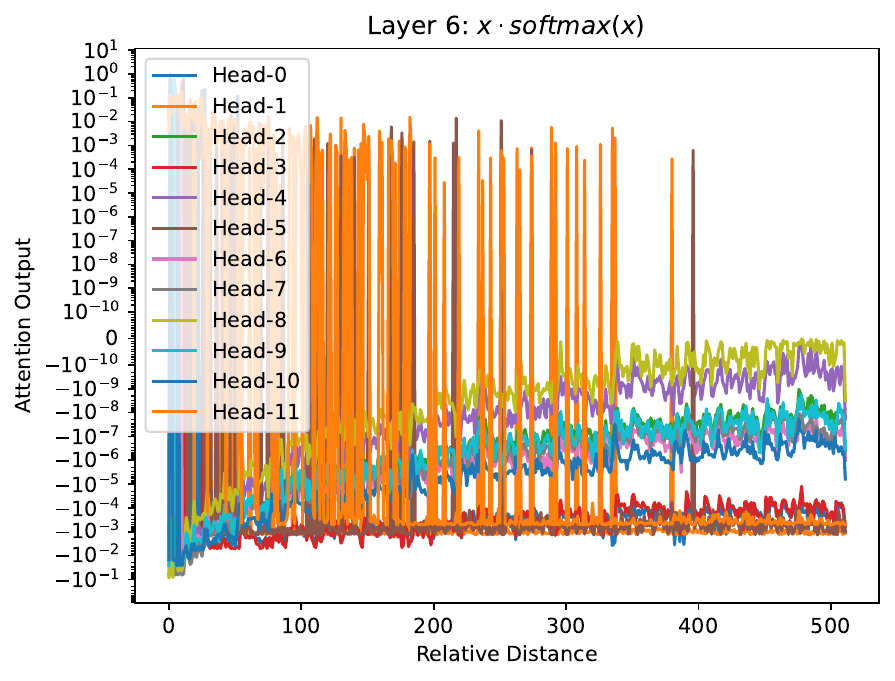}
\hspace{0in}
\includegraphics[width=0.32\textwidth]{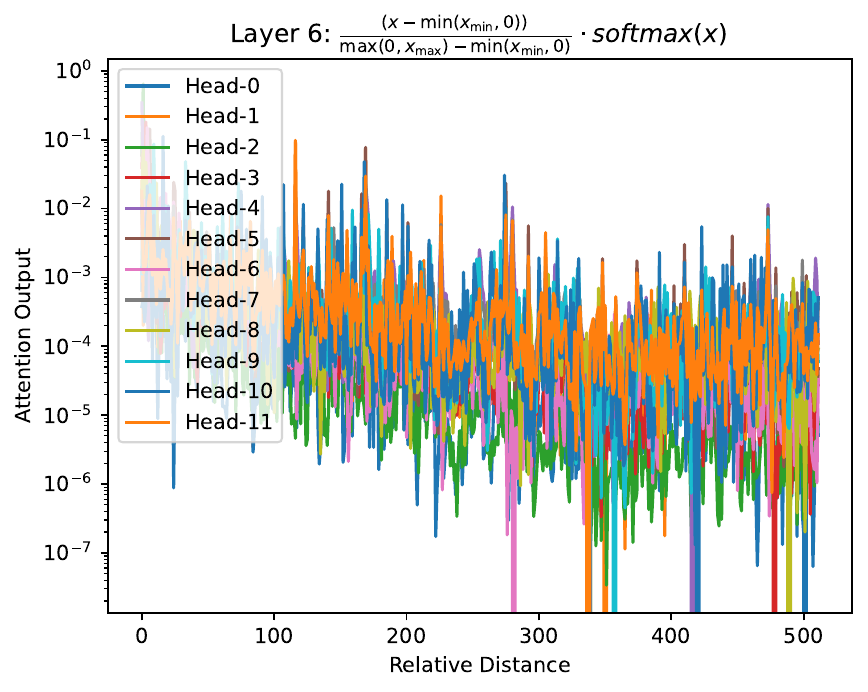}
\hspace{0in}

\includegraphics[width=0.32\textwidth]{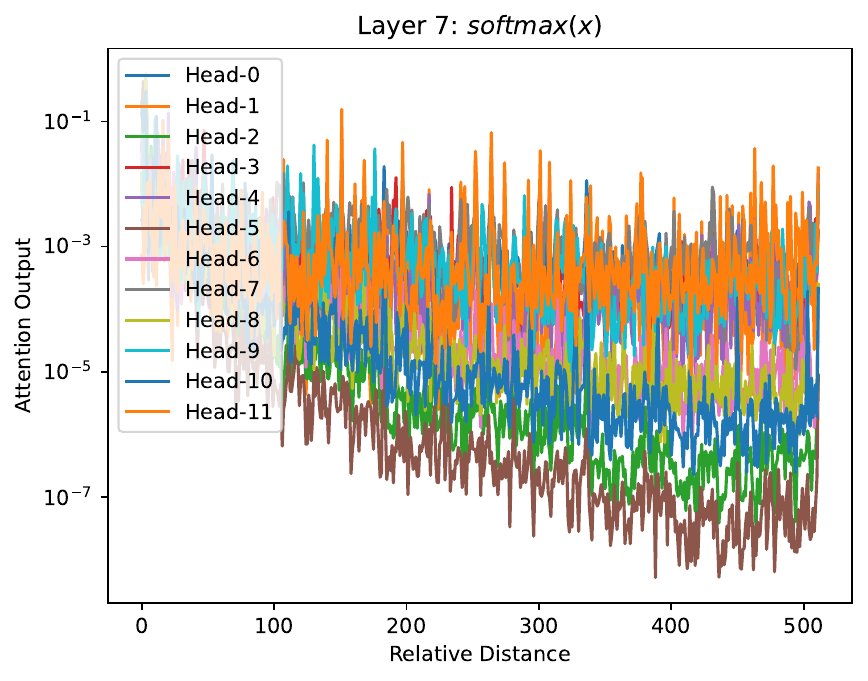}
\hspace{0in}
\includegraphics[width=0.32\textwidth]{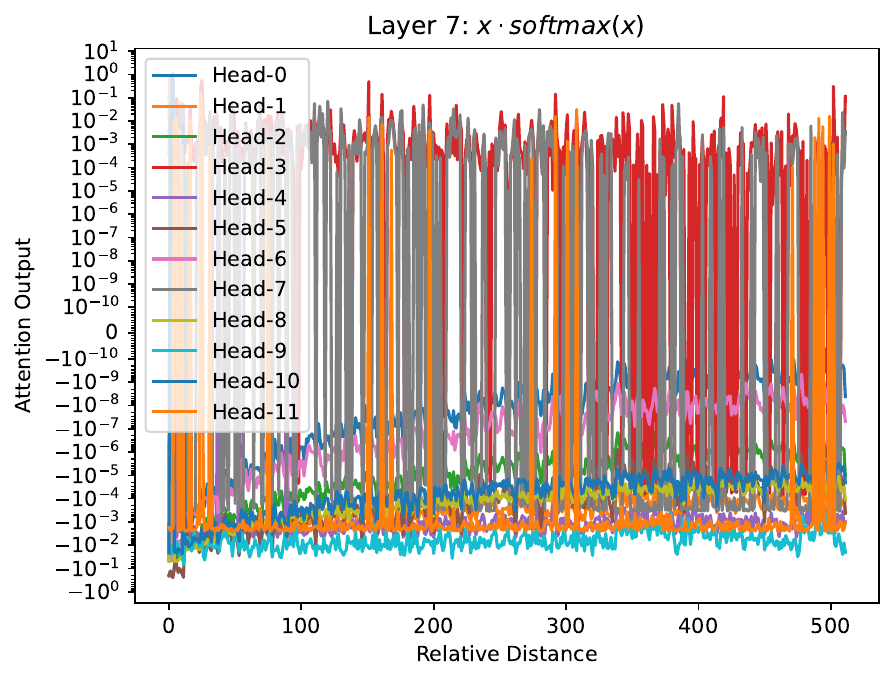}
\hspace{0in}
\includegraphics[width=0.32\textwidth]{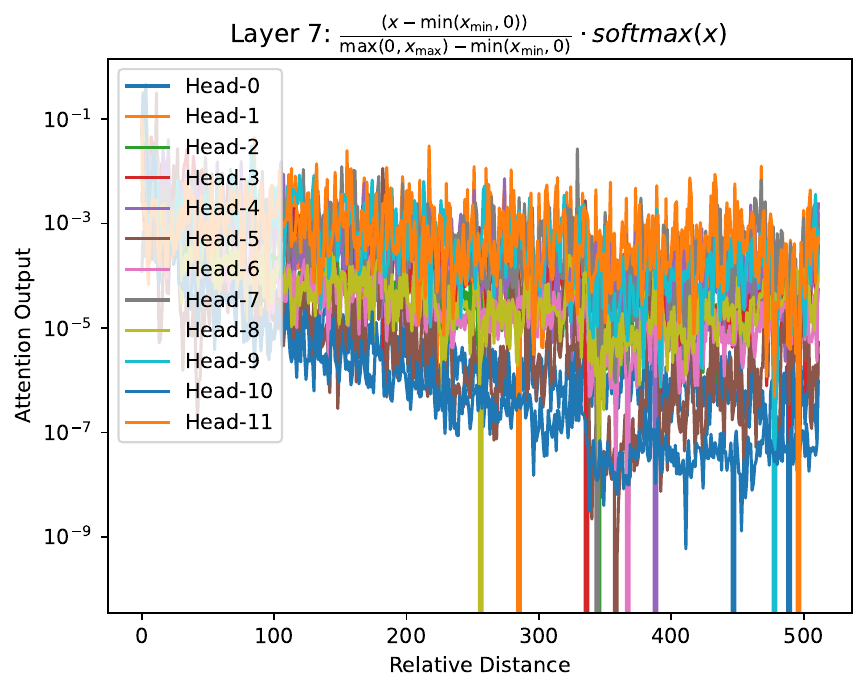}
\hspace{0in}

\includegraphics[width=0.32\textwidth]{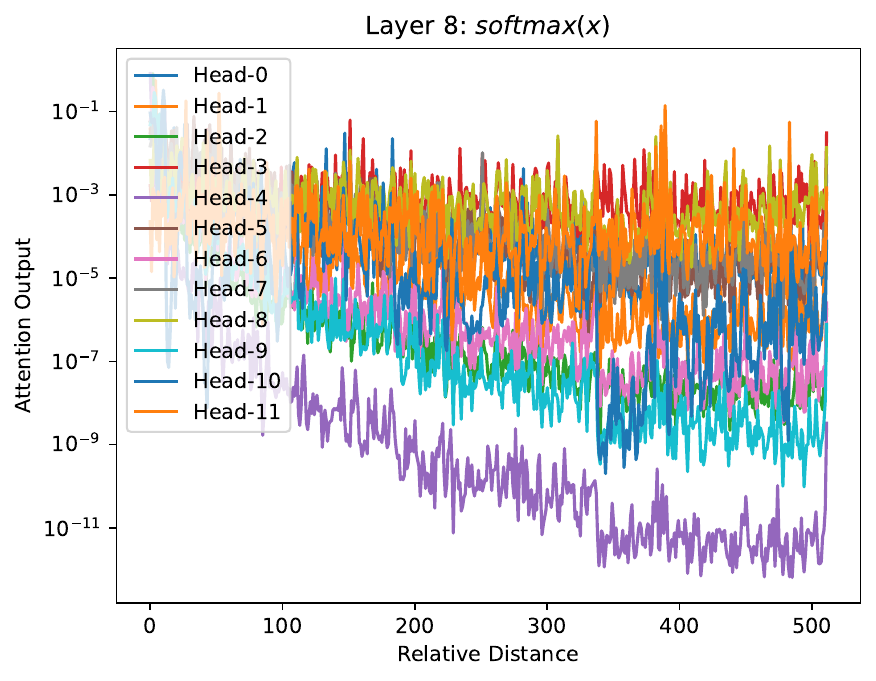}
\hspace{0in}
\includegraphics[width=0.32\textwidth]{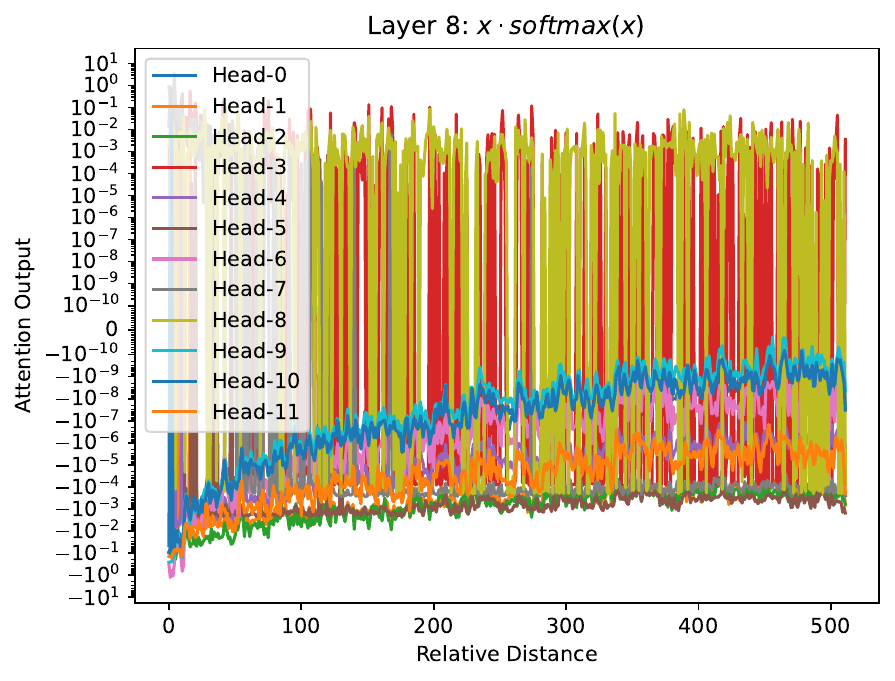}
\hspace{0in}
\includegraphics[width=0.32\textwidth]{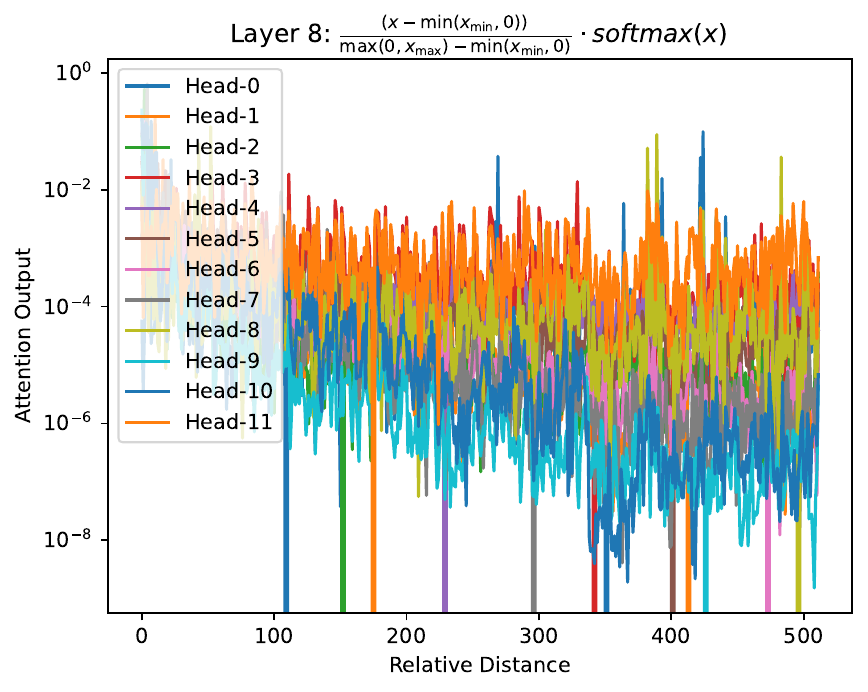}
\hspace{0in}

\includegraphics[width=0.32\textwidth]{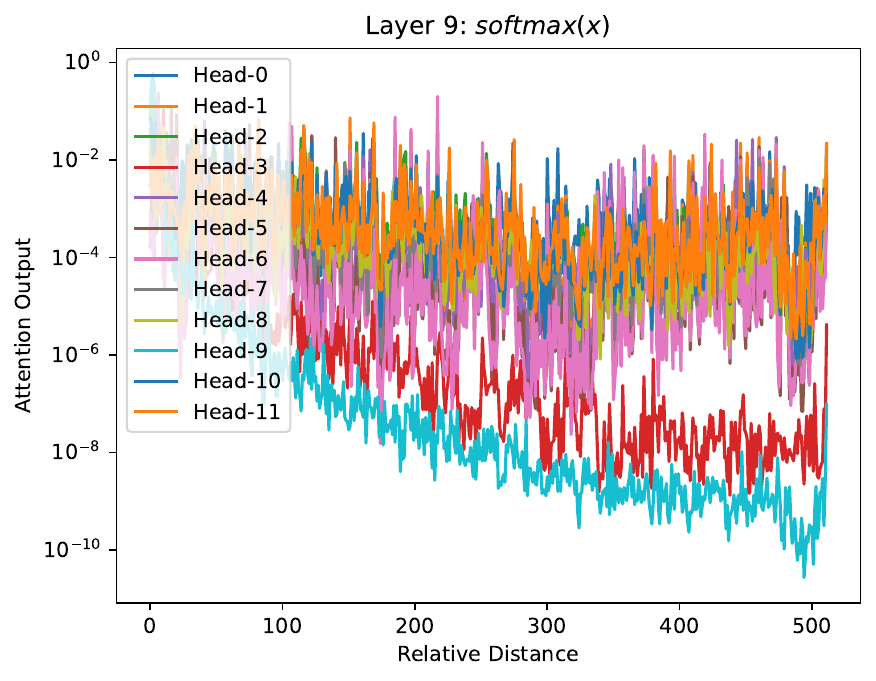}
\hspace{0in}
\includegraphics[width=0.32\textwidth]{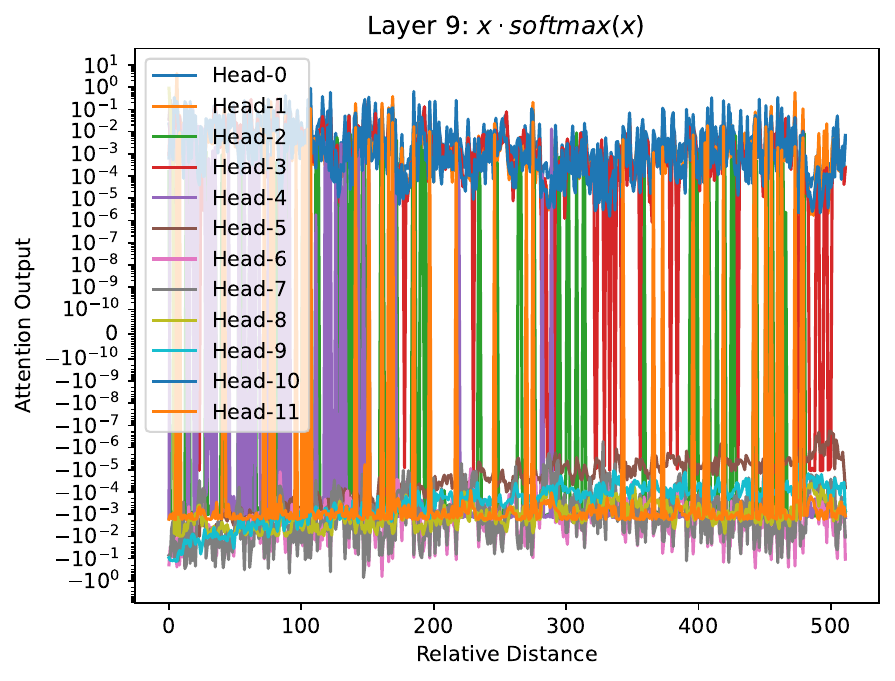}
\hspace{0in}
\includegraphics[width=0.32\textwidth]{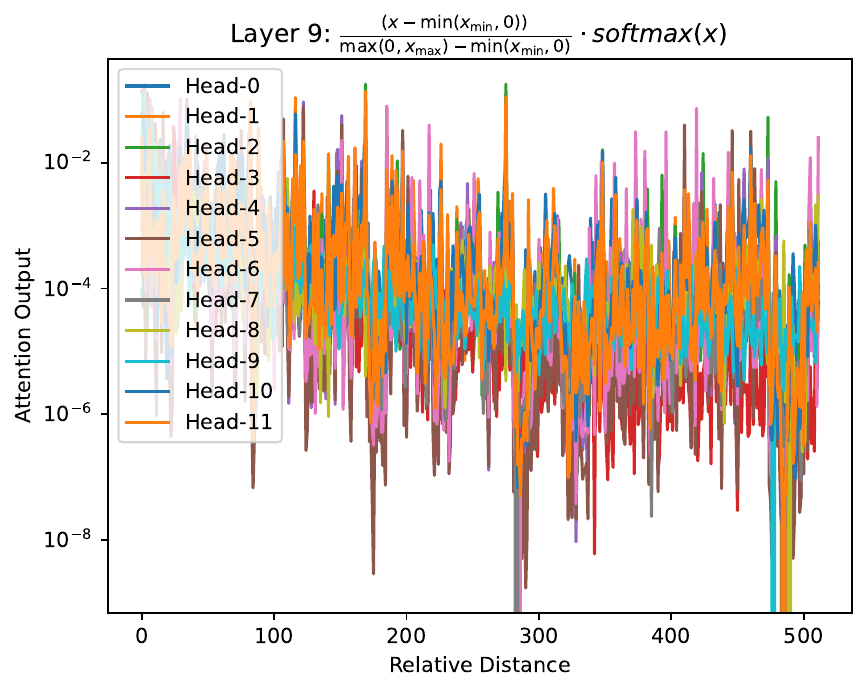}
\hspace{0in}

\setlength{\abovecaptionskip}{0.1cm}
\centering
\includegraphics[width=0.32\textwidth]{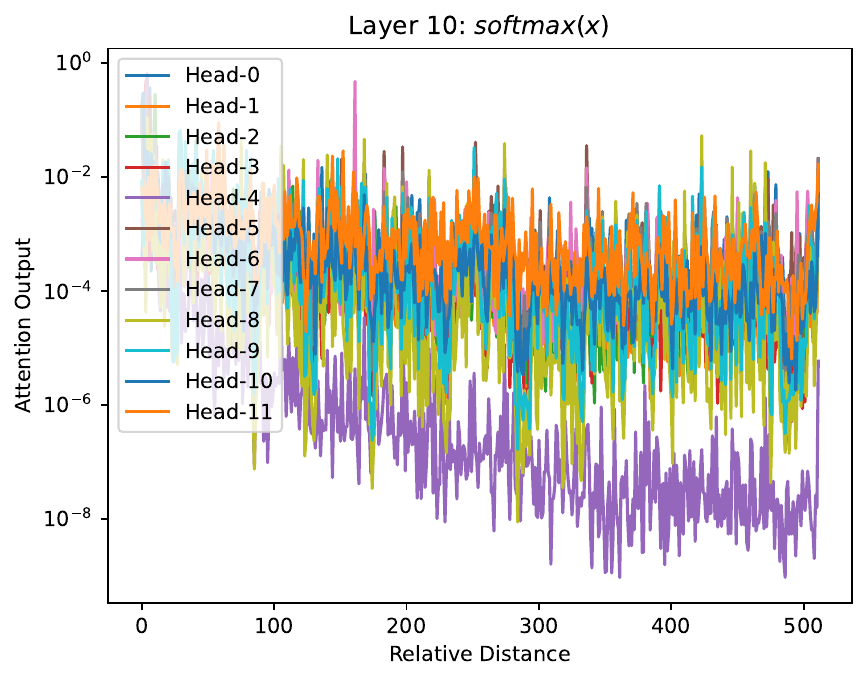}
\hspace{0in}
\includegraphics[width=0.32\textwidth]{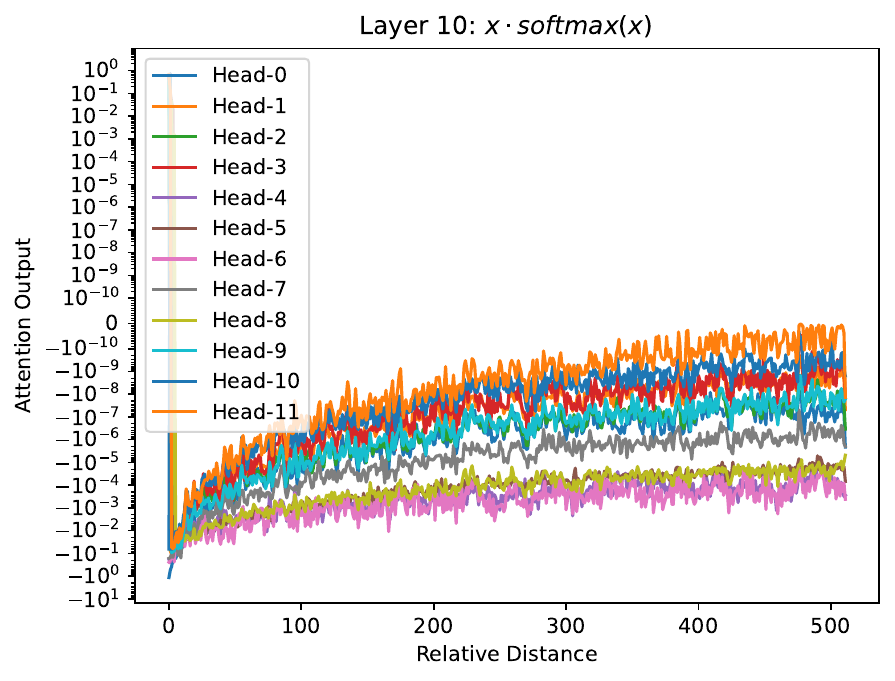}
\hspace{0in}
\includegraphics[width=0.32\textwidth]{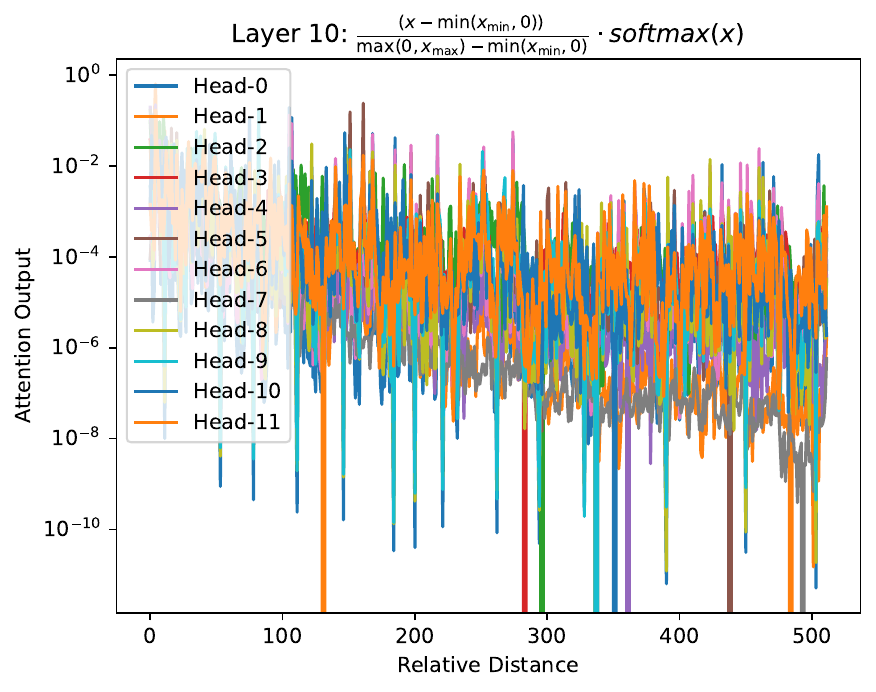}
\hspace{0in}

\caption{
\small
\textbf{The visualization of attention output, from left to right: 1) $softmax(x)$; 2) $x*softmax(x)$; 3) $\frac{(x - min(x_{\min},0))}{max(0,x_{max})-min(x_{min},0)} \cdot softmax(x)$.
}
}

\end{figure}

\clearpage
\begin{figure}[!htbp]




\includegraphics[width=0.32\textwidth]{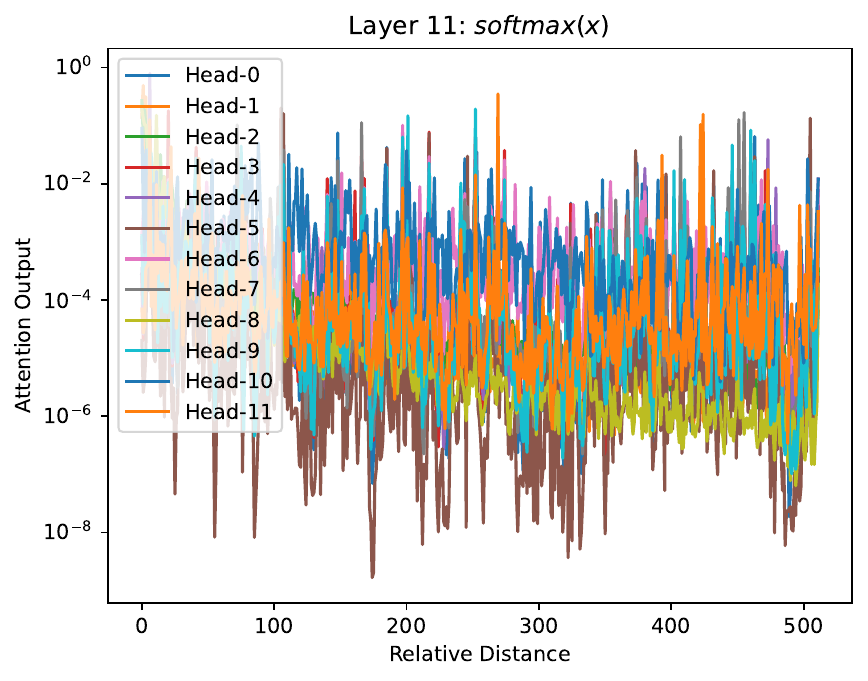}
\hspace{0in}
\includegraphics[width=0.32\textwidth]{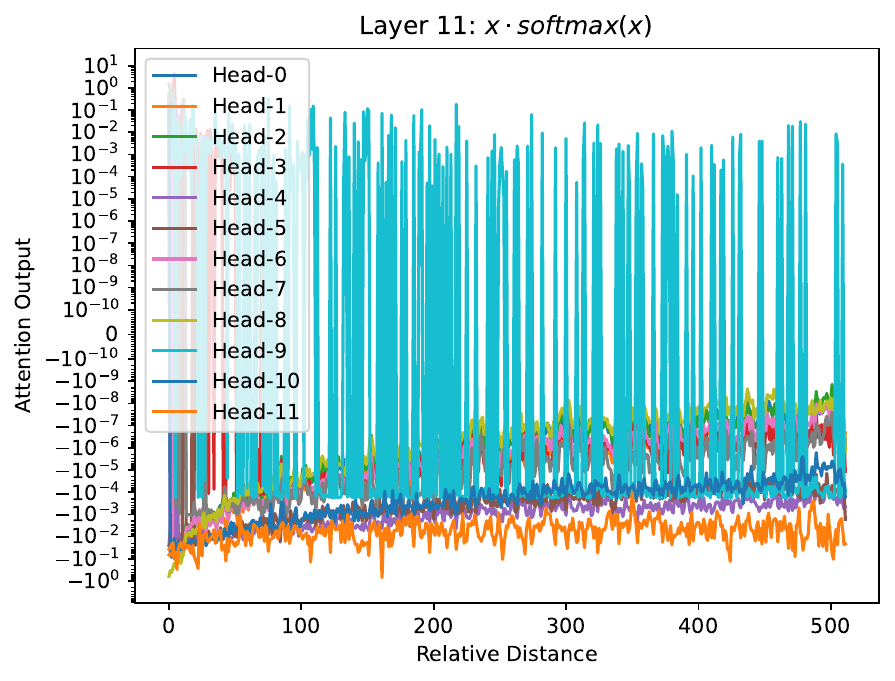}
\hspace{0in}
\includegraphics[width=0.32\textwidth]{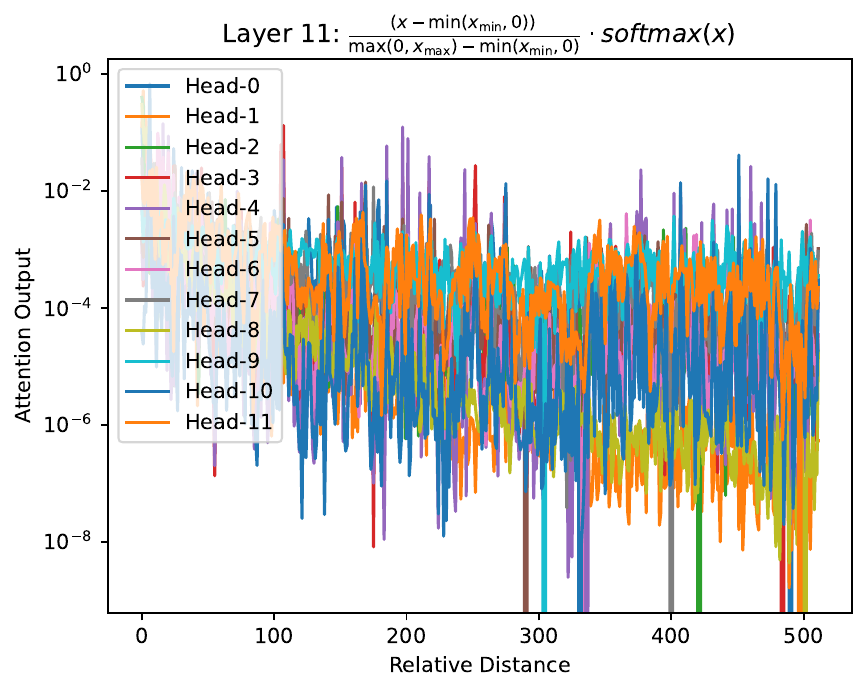}
\hspace{0in}

\includegraphics[width=0.32\textwidth]{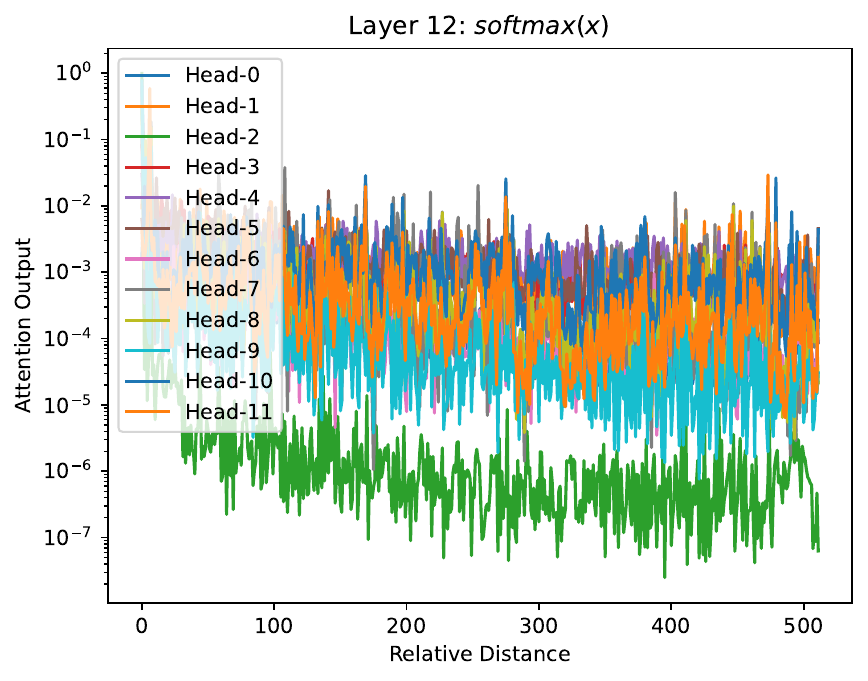}
\hspace{0in}
\includegraphics[width=0.32\textwidth]{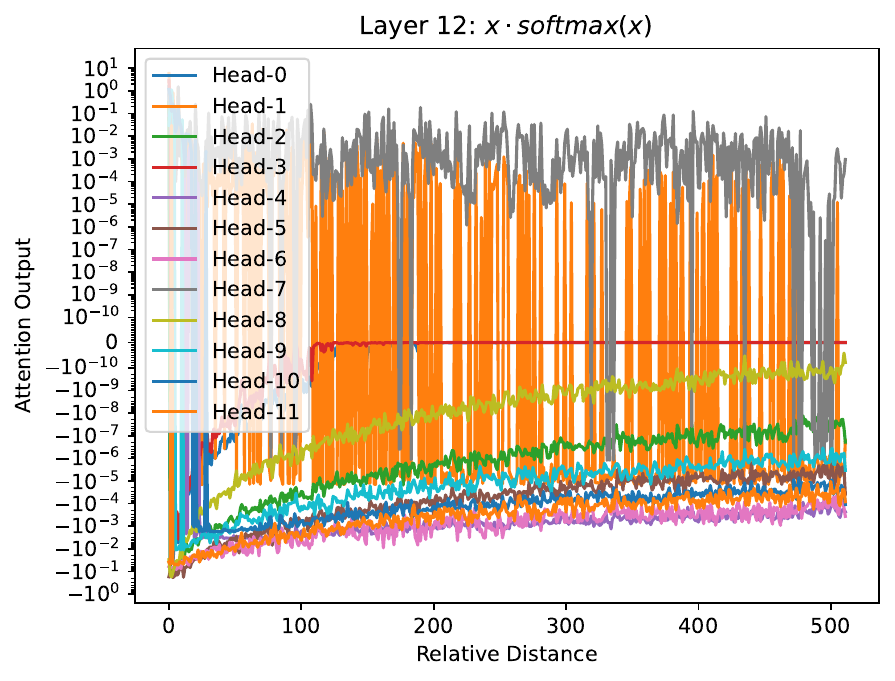}
\hspace{0in}
\includegraphics[width=0.32\textwidth]{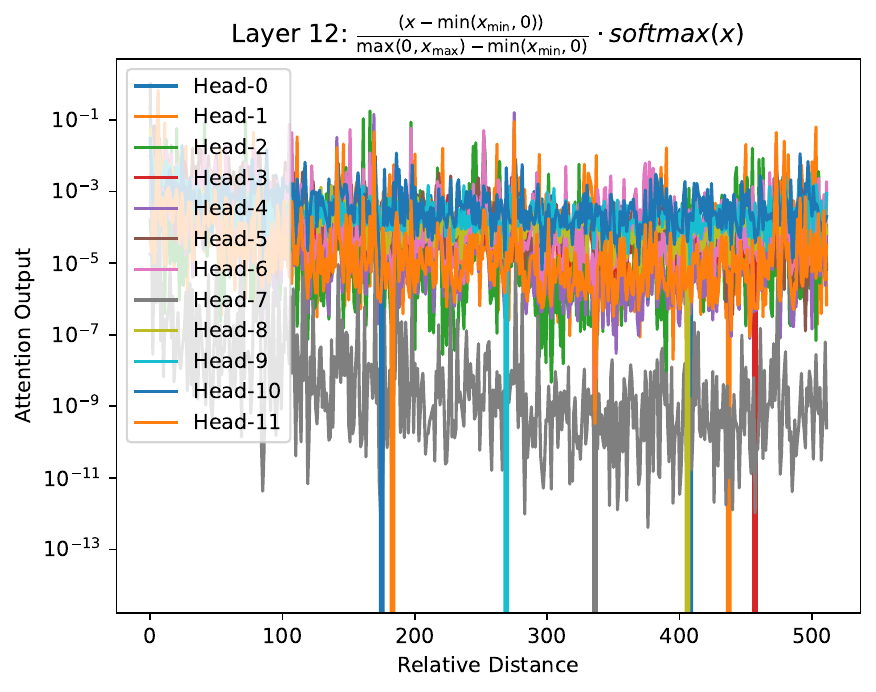}
\hspace{0in}

\caption{
\small
\textbf{The visualization of attention output, from left to right: 1) $softmax(x)$; 2) $x*softmax(x)$; 3) $\frac{(x - min(x_{\min},0))}{max(0,x_{max})-min(x_{min},0)} \cdot softmax(x)$.
}
}

\end{figure}

\clearpage
\newpage
\section{Implementation}
\label{appendix: implementation}
In this section, we present the implementation of the proposed \methodShortName module in \texttt{PyTorch} which allows for research purpose \citep{paszke2019pytorch}.

\definecolor{lightgreen}{rgb}{0,0.8,0}
\definecolor{darkgreen}{rgb}{0,0.8.0.2}
\definecolor{backcolour}{rgb}{0.97,0.97,0.94}
\lstset{language=Python,
basicstyle=\footnotesize,
breaklines=true,
backgroundcolor = \color{backcolour},
keywordstyle=\color{blue}\ttfamily,
stringstyle=\color{lightgreen}\ttfamily,
commentstyle=\color{gray}\ttfamily,
xleftmargin=2.5em,xrightmargin=0.5em, aboveskip=1em,
morecomment=[l][\color{darkgreen}]{\#}}

\begin{lstlisting}
import torch
import torch.nn as nn

class SA-Softmax(nn.Module):
  def __init__(self, operation_name):
    """


    Args:
      operation_name: "softmax","v1","v2","v3", or "v4"
    """
    super(SA-Softmax, self).__init__()


    self.operation_name= operation_name:
    

  def forward(self, attention: torch.Tensor, bias: torch.Tensor):
    """
    Args:
      attention: input sequence, which is q^T * k,
         shape [bsz, num_heads, seq_len, seq_len]
      bias: bias matrix, which can be generated by ALiBi, Kerple 
      FIRE or other additive position encodings
         shape [1,num_heads, seq_len, seq_len]

    Returns:
      attention with SA-Softmax,
      shape [bsz, num_heads, seq_len, seq_len]
    """
    attention_probs = softmax(attention_scores, attention_mask)
        
    if self.gradient_name=="v1":
        attention_probs=attention_probs*attention_scores
        attention_probs=torch.tril(attention_probs)
        
       
    elif self.gradient_name=="v2":
        B, H, T, _ = attention_scores.shape
        # Create a mask for the lower triangular part (including diagonal)
        mask = torch.tril(torch.ones(T, T, dtype=torch.bool, device= attention_scores.device))
        # Apply mask to get lower triangular values, replace upper triangle with inf (so it doesn't affect min)
        x_lower_tri = attention_scores.masked_fill(~mask, float('inf'))
        # Get the minimum value along the last dimension
        min_attention_score, _ = x_lower_tri.min(dim=-1,keepdim=True)
        attention_scores=torch.tril(attention_scores)
        attention_probs=attention_probs*(attention_scores-min_attention_score)
        attention_probs=torch.tril(attention_probs)

    elif self.gradient_name=="v3":
        B, H, T, _ = attention_scores.shape

        # Create a mask for the lower triangular part (including diagonal)
        mask = torch.tril(torch.ones(T, T, dtype=torch.bool, device= attention_scores.device))
        # Apply mask to get lower triangular values, replace upper triangle with inf (so it doesn't affect min)
        x_lower_tri = attention_scores.masked_fill(~mask, float('inf'))
        # Get the minimum value along the last dimension
        min_attention_score, _ = x_lower_tri.min(dim=-1,keepdim=True)
        # Apply mask to get lower triangular values, replace upper triangle with inf (so it doesn't affect min)
        x_lower_tri = attention_scores.masked_fill(~mask, float('-inf'))
        max_attention_score, _ = x_lower_tri.max(dim=-1,keepdim=True)

        attention_probs=attention_probs*((attention_scores-min_attention_score)/(max_attention_score-min_attention_score+1e-10))
    elif  self.gradient_name=="v4":
        attention_scores_tril_this=torch.tril(attention_scores)
        min_attention_score=torch.min(attention_scores_tril_this, -1,keepdim=True)[0]
        max_attention_score=torch.max(attention_scores_tril_this, -1,keepdim=True)[0]
        min_attention_score=torch.minimum(min_attention_score, torch.zeros(1, device= attention_scores.device))
        max_attention_score=torch.maximum(max_attention_score,torch.zeros(1, device= attention_scores.device))
        attention_probs=attention_probs*((attention_scores-min_attention_score)/(max_attention_score-min_attention_score+1e-10))
    
`   attention_probs=torch.tril(attention_probs)
    return attention_probs
\end{lstlisting}

\end{document}